\documentclass[10pt,twocolumn,letterpaper]{article}

\usepackage{cvpr}
\usepackage{times}
\usepackage{epsfig}
\usepackage{graphicx}
\usepackage{amsmath}
\usepackage{amssymb}

\usepackage{color,xcolor}
\usepackage{epsfig}
\usepackage{graphicx}

\usepackage{adjustbox}
\usepackage{array}
\usepackage{booktabs}
\usepackage{colortbl}
\usepackage{float,wrapfig}
\usepackage{hhline}
\usepackage{multirow}
\usepackage{subcaption} 
\usepackage[font={small}]{caption}

\usepackage{amsmath,amsfonts,amsthm,amssymb}
\usepackage{bm}
\usepackage{nicefrac}
\usepackage{microtype}

\usepackage{changepage}
\usepackage{extramarks}
\usepackage{fancyhdr}
\usepackage{lastpage}
\usepackage{setspace}
\usepackage{soul}
\usepackage{xspace}

\usepackage{url}

\usepackage{algorithm, algorithmic}
\usepackage{enumerate}

\usepackage{soul}
\usepackage{nth}

\newcolumntype{L}[1]{>{\raggedright\let\newline\\\arraybackslash\hspace{0pt}}m{#1}}
\newcolumntype{C}[1]{>{\centering\let\newline\\\arraybackslash\hspace{0pt}}m{#1}}
\newcolumntype{R}[1]{>{\raggedleft\let\newline\\\arraybackslash\hspace{0pt}}m{#1}}

\newcommand{\eqn}[1]{Equation~\ref{#1}}

\newcommand{\fig}[1]{Figure~\ref{#1}}
\newcommand{\figs}[1]{Figures~\ref{#1}}
\newcommand{\tbl}[1]{Table~\ref{#1}}

\newcommand{\ignore}[1]{}
\newcommand{\norm}[1]{\lVert#1\rVert}

\makeatletter
\DeclareRobustCommand\onedot{\futurelet\@let@token\@onedot}
\def\@onedot{\ifx\@let@token.\else.\null\fi\xspace}

\def\ie{\emph{i.e}\onedot} 
 
\def\etc{\emph{etc}\onedot} 
 
\def\etal{\emph{et al}\onedot}
\def\X{\mathbf{X}}
\def\x{\mathbf{x}}
\makeatother

\definecolor{MyDarkBlue}{rgb}{0,0.08,1}
\definecolor{MyDarkGreen}{rgb}{0.02,0.6,0.02}
\definecolor{MyDarkRed}{rgb}{0.8,0.02,0.02}
\definecolor{MyDarkOrange}{rgb}{0.40,0.2,0.02}
\definecolor{MyPurple}{RGB}{111,0,255}
\definecolor{MyRed}{rgb}{1.0,0.0,0.0}
\definecolor{MyGold}{rgb}{0.75,0.6,0.12}
\definecolor{MyDarkgray}{rgb}{0.66, 0.66, 0.66}

\def\xdd{\bm{X}_{\text{2D}}}
\def\xddd{\bm{X}_{\text{3D}}}

\def\cL{\mathcal{L}}

\newcommand{\myparagraph}[1]{\vspace{-14pt}\paragraph{#1}}

\newcommand{\data}{Pix3D\xspace}

\usepackage[pagebackref=true,breaklinks=true,colorlinks,bookmarks=false]{hyperref}

\cvprfinalcopy 

\ifcvprfinal\fi
\begin{document}

\title{\data: Dataset and Methods for Single-Image 3D Shape Modeling}

\author{Xingyuan Sun$^{*1,2}$ $\;\;$ Jiajun Wu$^{*1}$ $\;\;$ Xiuming Zhang$^1$ $\;\;$ Zhoutong Zhang$^1$\\Chengkai Zhang$^1$ $\;\;$ Tianfan Xue$^3$ $\;\;$ Joshua B. Tenenbaum$^1$ $\;\;$ William T. Freeman$^{1,3}$\\
\ \\
$^1$Massachusetts Institute of Technology $\;\;$ $^2$Shanghai Jiao Tong University $\;\;$ $^3$Google Research}


\twocolumn[{%
\renewcommand\twocolumn[1][]{#1}%
\maketitle
\thispagestyle{empty}
    \centering
    \vspace{-5pt}
    \includegraphics[width=\linewidth]{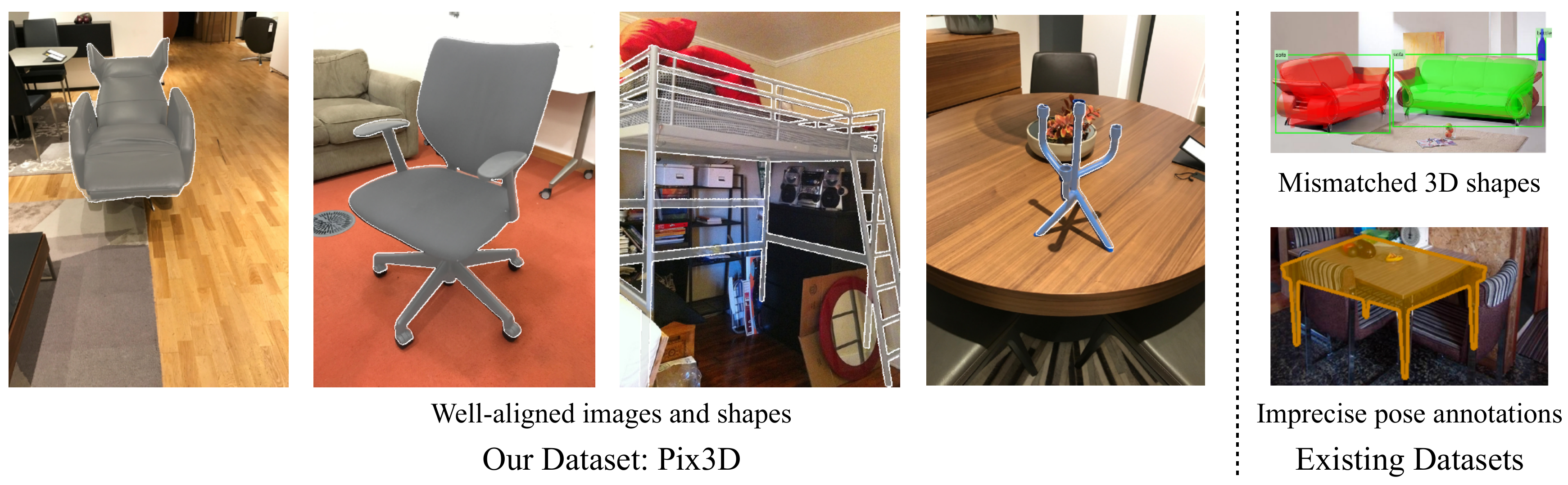}
    \vspace{-20pt}
    \captionof{figure}{
    We present \data, a new large-scale dataset of diverse image-shape pairs. Each 3D shape in \data is associated with a rich and diverse set of images, each with an accurate 3D pose annotation to ensure precise 2D-3D alignment. In comparison, existing datasets have limitations: 3D models may not match the objects in images; pose annotations may be imprecise; or the dataset may be relatively small.
    }
    \label{fig:teaser} 
    \vspace{1.2em}
}]

\footnotetext{$*$ indicates equal contributions.}
\begin{abstract}

\vspace{-5pt}

We study 3D shape modeling from a single image and make contributions to it in three aspects. First, we present \data, a large-scale benchmark of diverse image-shape pairs with pixel-level 2D-3D alignment. \data has wide applications in shape-related tasks including reconstruction, retrieval, viewpoint estimation, \etc. Building such a large-scale dataset, however, is highly challenging; existing datasets either contain only synthetic data, or lack precise alignment between 2D images and 3D shapes, or only have a small number of images. Second, we calibrate the evaluation criteria for 3D shape reconstruction through behavioral studies, and use them to objectively and systematically benchmark cutting-edge reconstruction algorithms on \data. Third, we design a novel model that simultaneously performs 3D reconstruction and pose estimation; our multi-task learning approach achieves state-of-the-art performance on both tasks. 

\vspace{-5pt}

\end{abstract}

\section{Introduction}
\label{sec:intro}

The computer vision community has put major efforts in building datasets. In 3D vision, there are rich 3D CAD model repositories like ShapeNet~\cite{Chang2015Shapenet:} and the Princeton Shape Benchmark~\cite{Shilane2004princeton}, large-scale datasets associating images and shapes like Pascal 3D+~\cite{Xiang2014PASCAL:} and ObjectNet3D~\cite{Xiang2016Objectnet3d:}, and benchmarks with fine-grained pose annotations for shapes in images like IKEA~\cite{Lim2013Parsing}. Why do we need one more?

Looking into \fig{fig:teaser}, we realize existing datasets have limitations for the task of modeling a 3D object from a single image. ShapeNet is a large dataset for 3D models, but does not come with real images; Pascal 3D+ and ObjectNet3D have real images, but the image-shape alignment is rough because the 3D models do not match the objects in images; IKEA has high-quality image-3D alignment, but it only contains 90 3D models and 759 images.

We desire a dataset that has all three merits---a large-scale dataset of real images and ground-truth shapes with precise 2D-3D alignment. Our dataset, named \data, has 395 3D shapes of nine object categories. Each shape associates with a set of real images, capturing the exact object in diverse environments. Further, the 10,069 image-shape pairs have precise 3D annotations, giving pixel-level alignment between shapes and their silhouettes in the images.

Building such a dataset, however, is highly challenging. For each object, it is difficult to simultaneously collect its high-quality geometry and in-the-wild images. We can crawl many images of real-world objects, but we do not have access to their shapes; 3D CAD repositories offer object geometry, but do not come with real images. Further, for each image-shape pair, we need a precise pose annotation that aligns the shape with its projection in the image. 

We overcome these challenges by constructing \data in three steps. 
First, we collect a large number of image-shape pairs by crawling the web and performing 3D scans ourselves. Second, we collect 2D keypoint annotations of objects in the images on Amazon Mechanical Turk, with which we optimize for 3D poses that align shapes with image silhouettes. Third, we filter out image-shape pairs with a poor alignment and, at the same time, collect attributes (\ie, truncation, occlusion) for each instance, again by crowdsourcing.

In addition to high-quality data, we need a proper metric to objectively evaluate the reconstruction results. A well-designed metric should reflect the visual appealingness of the reconstructions. In this paper, we calibrate commonly used metrics, including intersection over union, Chamfer distance, and earth mover's distance, on how well they capture human perception of shape similarity. Based on this, we benchmark state-of-the-art algorithms for 3D object modeling on \data to demonstrate their strengths and weaknesses.

With its high-quality alignment, Pix3D is also suitable for object pose estimation and shape retrieval. To demonstrate that, we propose a novel model that performs shape and pose estimation simultaneously. Given a single RGB image, our model first predicts its 2.5D sketches, and then regresses the 3D shape and the camera parameters from the estimated 2.5D sketches. Experiments show that multi-task learning helps to boost the model's performance. 

Our contributions are three-fold. First, we build a new dataset for single-image 3D object modeling; \data has a diverse collection of image-shape pairs with precise 2D-3D alignment. Second, we calibrate metrics for 3D shape reconstruction based on their correlations with human perception, and benchmark state-of-the-art algorithms on 3D reconstruction, pose estimation, and shape retrieval. Third, we present a novel model that simultaneously estimates object shape and pose, achieving state-of-the-art performance on both tasks.
\section{Related Work}
\label{sec:related}

\paragraph{Datasets of 3D shapes and scenes.}

For decades, researchers have been building datasets of 3D objects, either as a repository of 3D CAD models~\cite{bogo2014faust,bronstein2008numerical,Shilane2004princeton} or as images of 3D shapes with pose annotations~\cite{Leibe2003Analyzing,Savarese20073D}. Both directions have witnessed the rapid development of web-scale databases: ShapeNet~\cite{Chang2015Shapenet:} was proposed as a large repository of more than 50K models covering 55 categories, and Xiang~\etal built Pascal 3D+~\cite{Xiang2014PASCAL:} and ObjectNet3D~\cite{Xiang2016Objectnet3d:}, two large-scale datasets with alignment between 2D images and the 3D shape inside. While these datasets have helped to advance the field of 3D shape modeling, they have their respective limitations: datasets like ShapeNet or Elastic2D3D~\cite{lahner2016efficient} do not have real images, and recent 3D reconstruction challenges using ShapeNet have to be exclusively on synthetic images~\cite{yi2017large}; Pascal 3D+ and ObjectNet3D have only rough alignment between images and shapes, because objects in the images are matched to a pre-defined set of CAD models, not their actual shapes. This has limited their usage as a benchmark for 3D shape reconstruction~\cite{Tulsiani2017Multi}. 

With depth sensors like Kinect~\cite{Izadi2011KinectFusion:,janoch2011category}, the community has built various RGB-D or depth-only datasets of objects and scenes. We refer readers to the review article from Firman~\cite{firman2016rgbd} for a comprehensive list. Among those, many object datasets are designed for benchmarking robot manipulation~\cite{calli2015benchmarking,hodan2017t,Lai2011large,singh2014bigbird}. These datasets often contain a relatively small set of hand-held objects in front of clean backgrounds. 
Tanks and Temples~\cite{knapitsch2017tanks} is an exciting new benchmark with 14 scenes, designed for high-quality, large-scale, multi-view 3D reconstruction. In comparison, our dataset, \data, focuses on reconstructing a 3D object from a single image, and contains much more real-world objects and images.

Probably the dataset closest to \data is the large collection of object scans from Choi~\etal~\cite{choi2016large}, which contains a rich and diverse set of shapes, each with an RGB-D video. Their dataset, however, is not ideal for single-image 3D shape modeling for two reasons. First, the object of interest may be truncated throughout the video; this is especially the case for large objects like sofas. 
Second, their dataset does not explore the various contexts that an object may appear in, as each shape is only associated with a single scan. In \data, we address both problems by leveraging powerful web search engines and crowdsourcing. 

Another closely related benchmark is IKEA~\cite{Lim2013Parsing}, which provides accurate alignment between images of IKEA objects and 3D CAD models. This dataset is therefore particularly suitable for fine pose estimation. However, it contains only 759 images and 90 shapes, relatively small for shape modeling\footnote{Only 90 of the 219 shapes in the IKEA dataset have associated images.}. In contrast, \data contains 10,069 images (13.3x) and 395 shapes (4.4x) of greater variations. 

Researchers have also explored constructing scene datasets with 3D annotations. Notable attempts include LabelMe-3D~\cite{russell2009building}, NYU-D~\cite{Silberman2012Indoor}, SUN RGB-D~\cite{song2015sun}, KITTI~\cite{geiger2012we}, and modern large-scale RGB-D scene datasets~\cite{dai2017scannet,McCormac2017SceneNet,Song2017Semantic}. These datasets are either synthetic or contain only 3D surfaces of real scenes. \data, in contrast, offers accurate alignment between 3D object shape and 2D images in the wild.

\myparagraph{Single-image 3D reconstruction.}
\begin{figure*}[t]
\centering
\includegraphics[width=\linewidth]{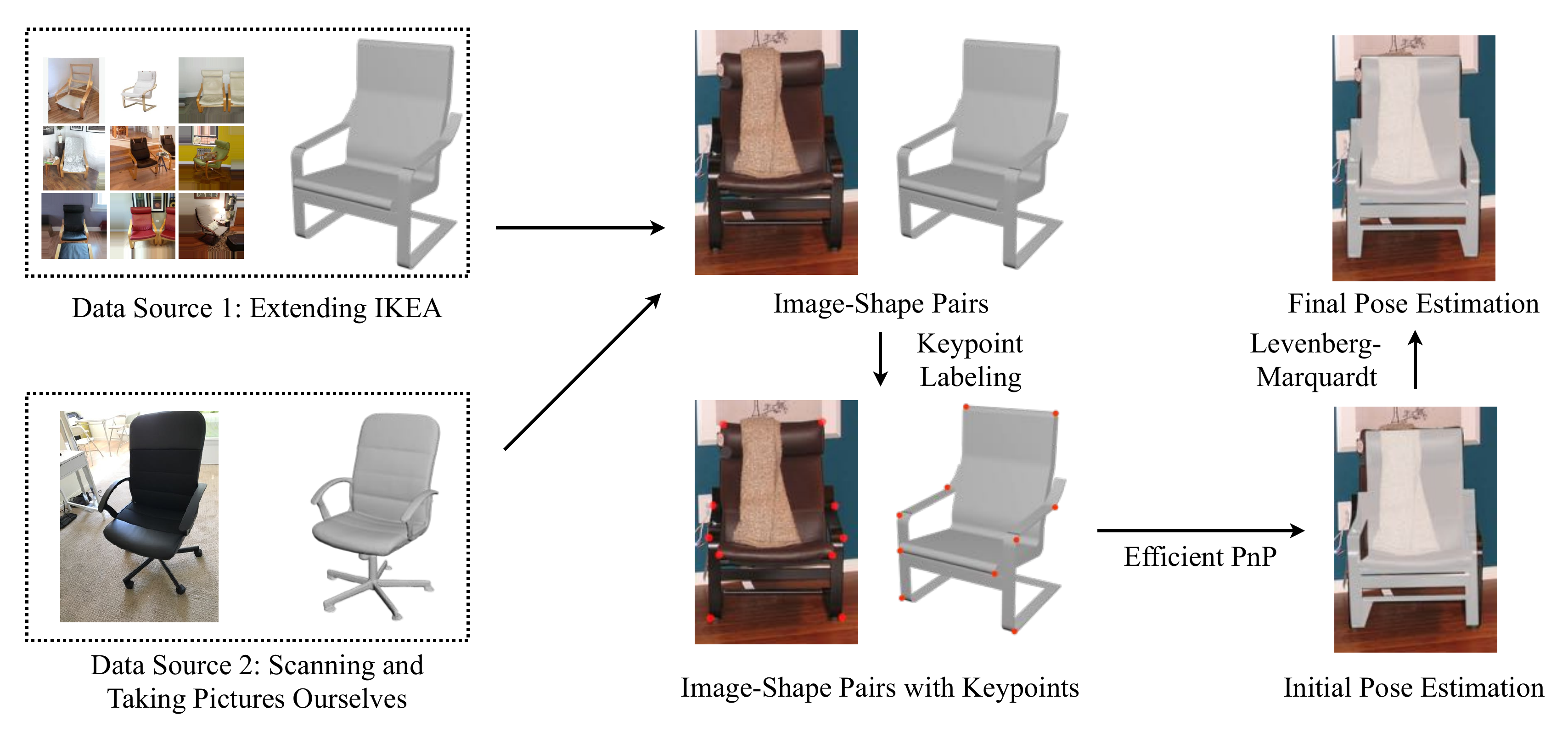}
\vspace{-23pt}
\caption{We build the dataset in two steps. First, we collect image-shape pairs by crawling web images of IKEA furniture as well as scanning objects and taking pictures ourselves. Second, we align the shapes with their 2D silhouettes by minimizing the 2D coordinates of the keypoints and their projected positions from 3D, using the Efficient PnP and the Levenberg-Marquardt algorithm.} 
\label{fig:data_pipeline}
\vspace{-15pt}
\end{figure*}

The problem of recovering object shape from a single image is challenging, as it requires both powerful recognition systems and prior shape knowledge. Using deep convolutional networks, researchers have made significant progress in recent years~\cite{Choy20163d,Girdhar2016Learning,hane2017hierarchical,Kar2015Category,Novotny2017Learning,Rezende2016Unsupervised,Tatarchenko2016Multi,Tulsiani2017Multi,marrnet,Wu2016Learning,Yan2016Perspective,Soltani2017Synthesizing,3dinterpreter}. While most of these approaches represent objects in voxels, there have also been attempts to reconstruct objects in point clouds~\cite{fan2017point} or octave trees~\cite{riegler2017octnet,tatarchenko2017octree}. In this paper, we demonstrate that our newly proposed \data serves as an ideal benchmark for evaluating these algorithms. We also propose a novel model that jointly estimates an object's shape and its 3D pose. 

\myparagraph{Shape retrieval.}

Another related research direction is retrieving similar 3D shapes given a single image, instead of reconstructing the object's actual geometry~\cite{Aubry2014Seeing,Fisher2010Context,gupta2015aligning,savva2016shrec}. \data contains shapes with significant inter-class and intra-class variations, and is therefore suitable for both general-purpose and fine-grained shape retrieval tasks.

\myparagraph{3D pose estimation.}

Many of the aforementioned object datasets include annotations of object poses~\cite{Leibe2003Analyzing,Lim2013Parsing,Savarese20073D,Xiang2016Objectnet3d:,Xiang2014PASCAL:}. Researchers have also proposed numerous methods on 3D pose estimation~\cite{Fidler20123D,ozuysal2009pose,Su2015Render,Tulsiani2015Viewpoints}. In this paper, we show that \data is also a proper benchmark for this task.
\section{Building \data}
\label{sec:data}

\fig{fig:data_pipeline} summarizes how we build \data. We collect raw images from web search engines and shapes from 3D repositories; we also take pictures and scan shapes ourselves. Finally, we use labeled keypoints on both 2D images and 3D shapes to align them. 

\subsection{Collecting Image-Shape Pairs}

We obtain raw image-shape pairs in two ways. One is to crawl images of IKEA furniture from the web and align them with CAD models provided in the IKEA dataset~\cite{Lim2013Parsing}. The other is to directly scan 3D shapes and take pictures. 

\myparagraph{Extending IKEA.}

The IKEA dataset~\cite{Lim2013Parsing} contains 219 high-quality 3D models of IKEA furniture, but has only 759 images for 90 shapes. Therefore, we choose to keep the 3D shapes from IKEA dataset, but expand the set of 2D images using online image search engines and crowdsourcing.

For each 3D shape, we first search for its corresponding 2D images through Google, Bing, and Baidu, using its IKEA model name as the keyword. We obtain 104,220 images for the 219 shapes. We then use Amazon Mechanical Turk (AMT) to remove irrelevant ones. For each image, we ask three AMT workers to label whether this image matches the 3D shape or not. For images whose three responses differ, we ask three additional workers and decide whether to keep them based on majority voting. We end up with 14,600 images for the 219 IKEA shapes.

\begin{figure*}[t]
\centering
\includegraphics[width=\linewidth]{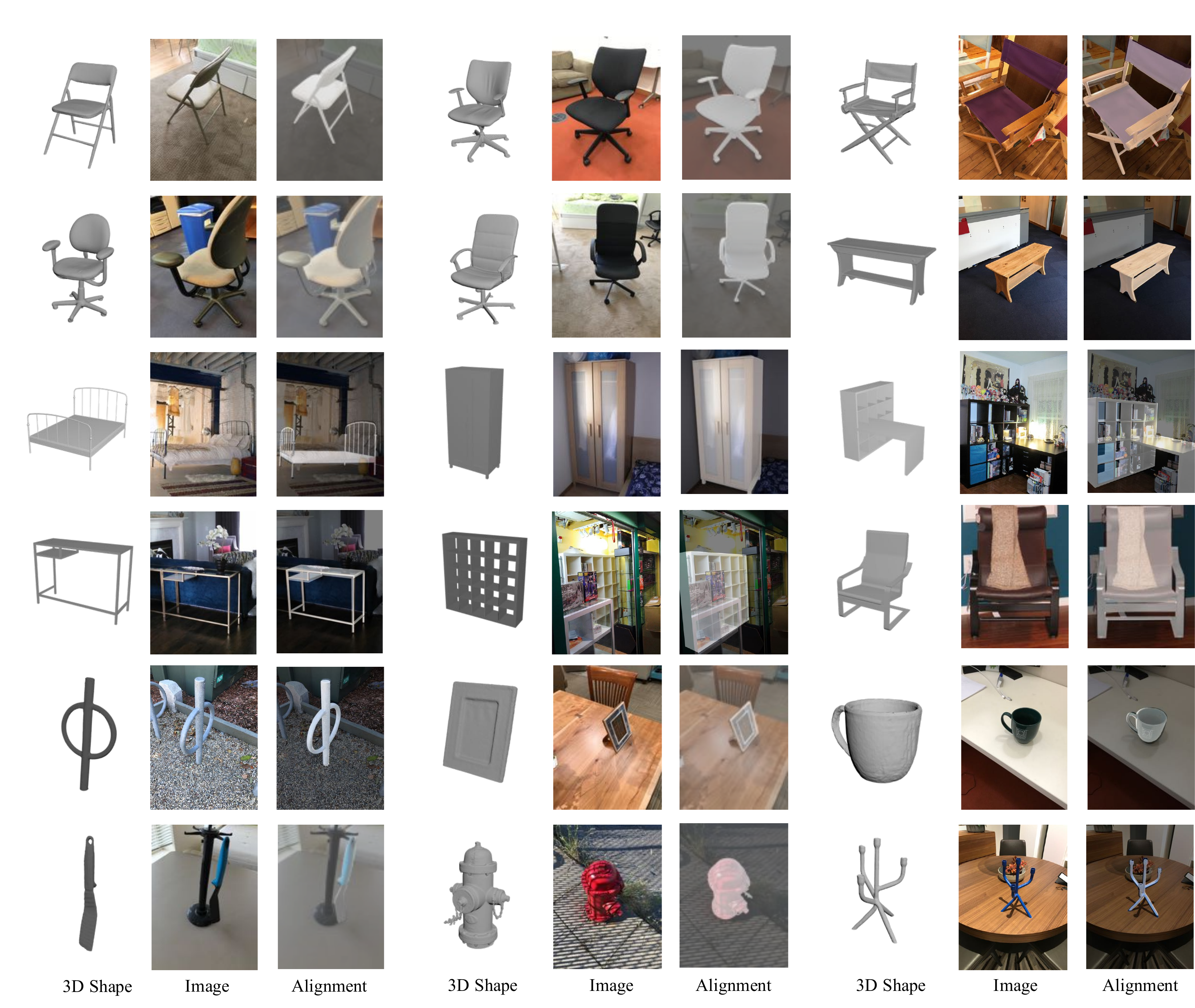}
\vspace{-20pt}
\caption{Sample images and shapes in \data. From left to right: 3D shapes, 2D images, and 2D-3D alignment. Rows 1--2 show some chairs we scanned, rows 3--4 show a few IKEA objects, and rows 5--6 show some objects of other categories we scanned.}
\vspace{-15pt}
\label{fig:data_sample}
\end{figure*}

\myparagraph{3D scan.}

We scan non-IKEA objects with a Structure Sensor\footnote{\url{https://structure.io}} mounted on an iPad. We choose to use the Structure Sensor because its mobility enables us to capture a wide range of shapes.

The iPad RGB camera is synchronized with the depth sensor at 30 Hz, and calibrated by the Scanner App provided by Occipital, Inc.\footnote{\url{https://occipital.com}} The resolution of RGB frames is 2592$\times$1936, and the resolution of depth frames is 320$\times$240. For each object, we take a short video and fuse the depth data to get its 3D mesh by using fusion algorithm provided by Occipital, Inc. We also take 10--20 images for each scanned object in front of various backgrounds from different viewpoints, making sure the object is neither cropped nor occluded. In total, we have scanned 209 objects and taken 2,313 images. Combining these with the IKEA shapes and images, we have 418 shapes and 16,913 images altogether.

\subsection{Image-Shape Alignment}

To align a 3D CAD model with its projection in a 2D image, we need to solve for its 3D pose (translation and rotation), and the camera parameters used to capture the image. 

We use a keypoint-based method inspired by Lim~\etal~\cite{Lim2013Parsing}. Denote the keypoints' 2D coordinates as $\xdd = \{\x_1, \x_2, \cdots, \x_n\}$ and their corresponding 3D coordinates as $\xddd = \{\X_1, \X_2, \cdots, \X_n\}$. We solve for camera parameters and 3D poses that minimize the reprojection error of the keypoints. Specifically, we want to find the projection matrix $\bm{P}$ that minimizes
\vspace{-5pt}
\begin{equation}
	\cL(\bm{P}; \xddd, \xdd) = \sum_{i} \norm{\text{Proj}_{\bm{P}}(\X_i) - \x_i}_2^2,\label{eq:obj_func}
\vspace{-5pt}
\end{equation} 
where $\text{Proj}_{\bm{P}}(\cdot)$ is the projection function. 

Under the central projection assumption (zero-skew, square pixel, and the optical center is at the center of the frame), we have $\bm{P}=\bm{K}[\bm{R}|\bm{T}]$, where $\bm{K}$ is the camera intrinsic matrix; $\bm{R} \in \mathbb{R}^{3\times 3}$ and $\bm{T} \in \mathbb{R}^3$ represent the object's 3D rotation and 3D translation, respectively. We know
\begin{equation}
    \bm{K} = 
    \begin{bmatrix}
        f & 0 & w / 2 \\
        0 & f & h / 2 \\
        0 & 0 & 1 \\
    \end{bmatrix}, 
\end{equation}
where $f$ is the focal length, and $w$ and $h$ are the width and height of the image. Therefore, there are altogether seven parameters to be estimated: rotations $\theta, \phi, \psi$, translations $x, y, z$, and focal length $f$ (Rotation matrix $R$ is determined by $\theta$, $\phi$, and $\psi$). 

To solve \eqn{eq:obj_func}, we first calculate a rough 3D pose using the Efficient P\textit{n}P algorithm~\cite{lepetit2009epnp} and then refine it using the Levenberg-Marquardt algorithm~\cite{levenberg1944method, marquardt1963algorithm}, as shown in \fig{fig:data_pipeline}. Details of each step are described below.

\myparagraph{Efficient P\textit{n}P.}

Perspective-\textit{n}-Point (P\textit{n}P) is the problem of estimating the pose of a calibrated camera given paired 3D points and 2D projections. The Efficient P\textit{n}P (EPnP) algorithm solves the problem using virtual control points~\cite{levenberg1944method}. Because EPnP does not estimate the focal length, we enumerate the focal length $f$ from 300 to 2,000 with a step size of 10, solve for the 3D pose with each $f$, and choose the one with the minimum projection error.

\myparagraph{The Levenberg-Marquardt algorithm (LMA).}

We take the output of EPnP with 50 random disturbances as the initial states, and run LMA on each of them. Finally, we choose the solution with the minimum projection error.

\myparagraph{Implementation details.}

For each 3D shape, we manually label its 3D keypoints. The number of keypoints ranges from 8 to 24. For each image, we ask three AMT workers to label if each keypoint is visible on the image, and if so, where it is. We only consider visible keypoints during the optimization.

The 2D keypoint annotations are noisy, which severely hurts the performance of the optimization algorithm. We try two methods to increase its robustness. The first is to use RANSAC. The second is to use only a subset of 2D keypoint annotations. For each image, denote $C=\{c_1, c_2, c_3\}$ as its three sets of human annotations. We then enumerate the seven nonempty subsets $C_k\subseteq C$; for each keypoint, we compute the median of its 2D coordinates in $C_k$. We apply our optimization algorithm on every subset $C_k$, and keep the output with the minimum projection error. After that, we let three AMT workers choose, for each image, which of the two methods offers better alignment, or neither performs well. At the same time, we also collect attributes (\ie, truncation, occlusion) for each image. Finally, we fine-tune the annotations ourselves using the GUI offered in ObjectNet3D~\cite{Xiang2016Objectnet3d:}. Altogether there are 395 3D shapes and 10,069 images. Sample 2D-3D pairs are shown in \fig{fig:data_sample}.

\begin{figure}[t!]
\centering
\includegraphics[width=\linewidth]{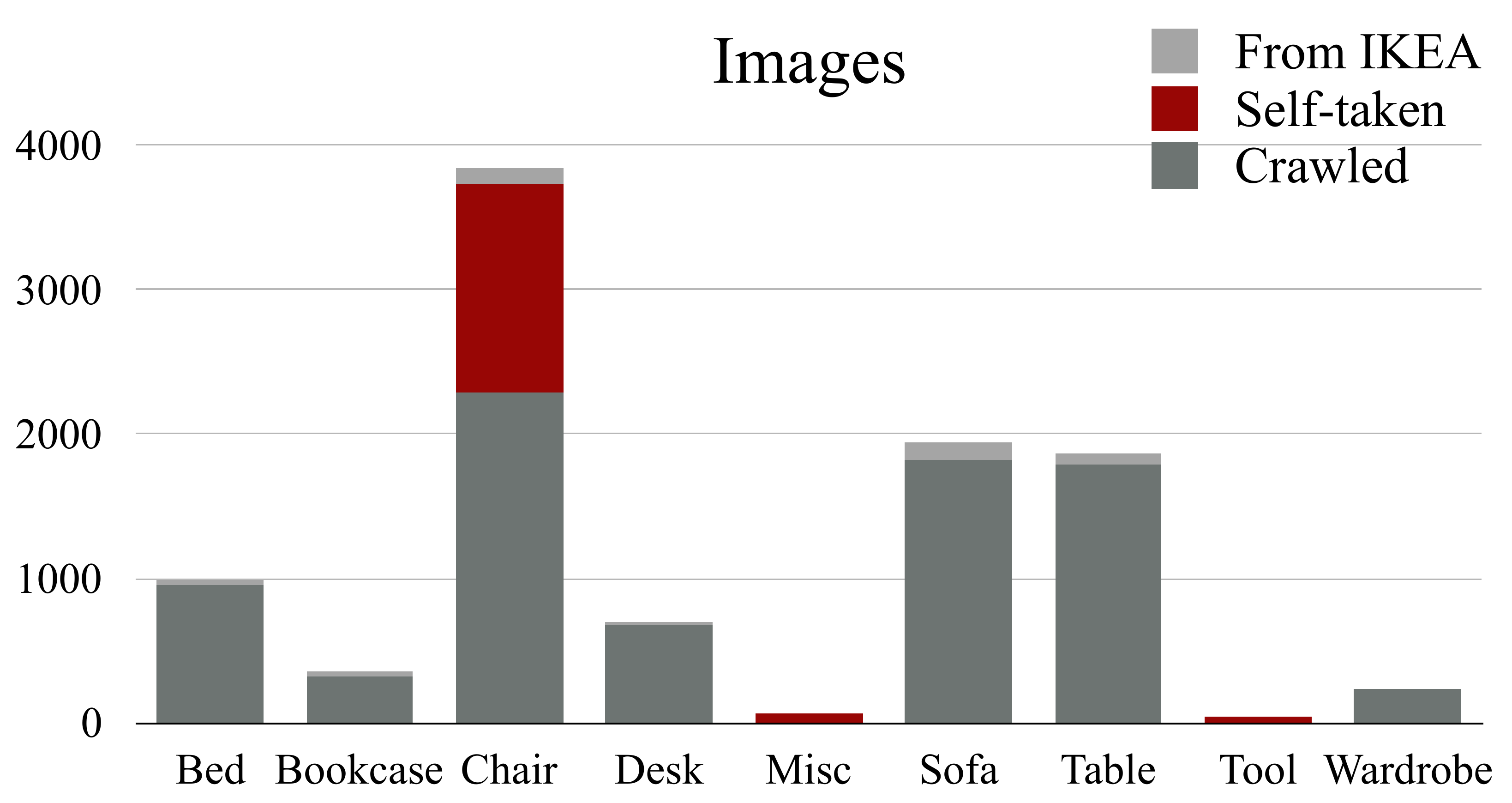}
\vspace{-20pt}
\caption{The distribution of images across categories}
\vspace{-20pt}
\label{fig:distr_img}
\end{figure}

\section{Exploring \data}
\label{sec:analysis}

We now present some statistics of \data, and contrast it with its predecessors. 

\myparagraph{Dataset statistics.}

\begin{figure}[t!]
\includegraphics[width=\linewidth]{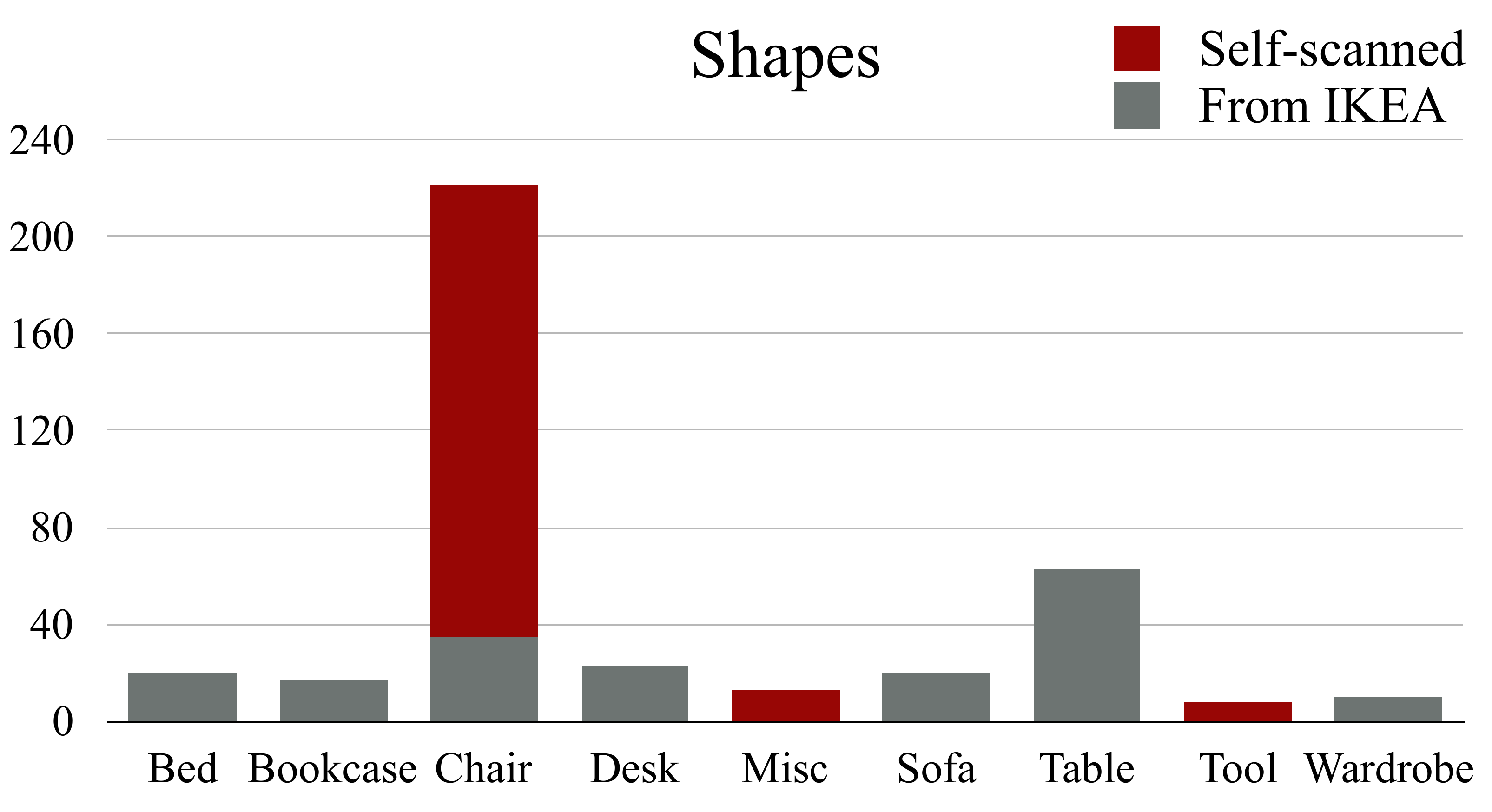}
\vspace{-20pt}
\caption{The distribution of shapes across categories}
\vspace{-10pt}
\label{fig:distr_model}
\end{figure}

\begin{figure}[t]
\centering
\includegraphics[width=\linewidth]{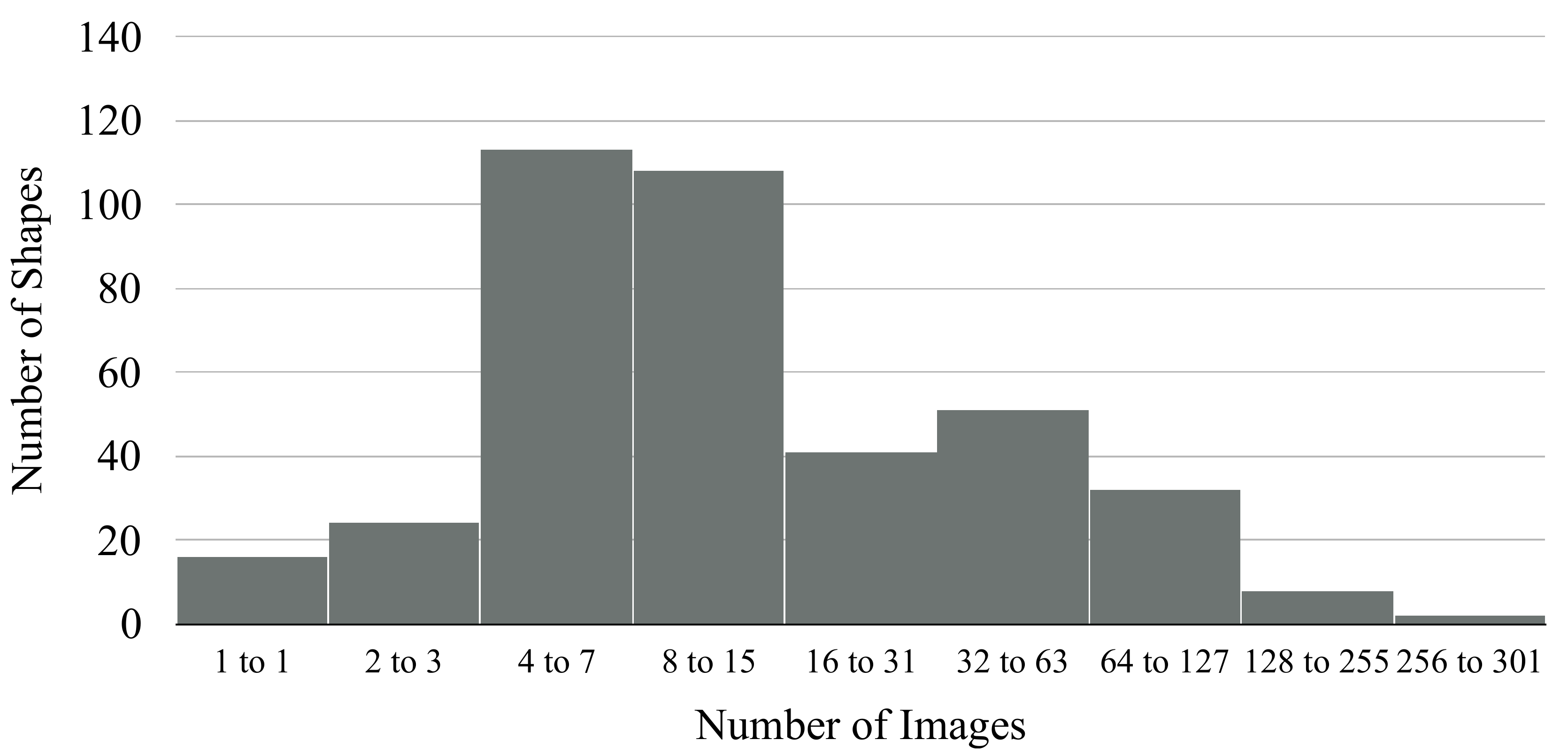}
\vspace{-20pt}
\caption{Number of images available for each shape}
\vspace{-15pt}
\label{fig:distr_num}
\end{figure}

\figs{fig:distr_img} and \ref{fig:distr_model} show the category distributions of 2D images and 3D shapes in \data; \fig{fig:distr_num} shows the distribution of the number of images each model has. Our dataset covers a large variety of shapes, each of which has a large number of in-the-wild images. Chairs cover the significant part of \data, because they are common, highly diverse, and well-studied by recent literature~\cite{Dosovitskiy2016Learning,Tulsiani2017Multi,gwak2017weakly}. 
\myparagraph{Quantitative evaluation.}

As a quantitative comparison on the quality of \data and other datasets, we randomly select 25 chair and 25 sofa images from PASCAL 3D+~\cite{Xiang2014PASCAL:}, ObjectNet3D~\cite{Xiang2016Objectnet3d:}, IKEA~\cite{Lim2013Parsing}, and \data. For each image, we render the projected 2D silhouette of the shape using its pose annotation provided by the dataset. We then manually annotate the ground truth object masks in these images, and calculate Intersection over Union (IoU) between the projections and the ground truth. For each image-shape pair, we also ask 50 AMT workers whether they think the image is picturing the 3D \emph{ground truth} shape provided by the dataset.

From \tbl{tbl:analy_iou}, we see that \data has much higher IoUs than PASCAL 3D+ and ObjectNet3D, and slightly higher IoUs compared with the IKEA dataset. Humans also feel IKEA and \data have matched images and shapes, but not PASCAL 3D+ or ObjectNet3D. In addition, we observe that many CAD models in the IKEA dataset are of an incorrect scale, making it challenging to align the shapes with images. For example, there are only 15 unoccluded and untruncated images of sofas in IKEA, while \data has 1,092.

\begin{figure*}[t]
\begin{minipage}[c]{0.33\linewidth}
    \centering
    \includegraphics[width=0.99\linewidth]{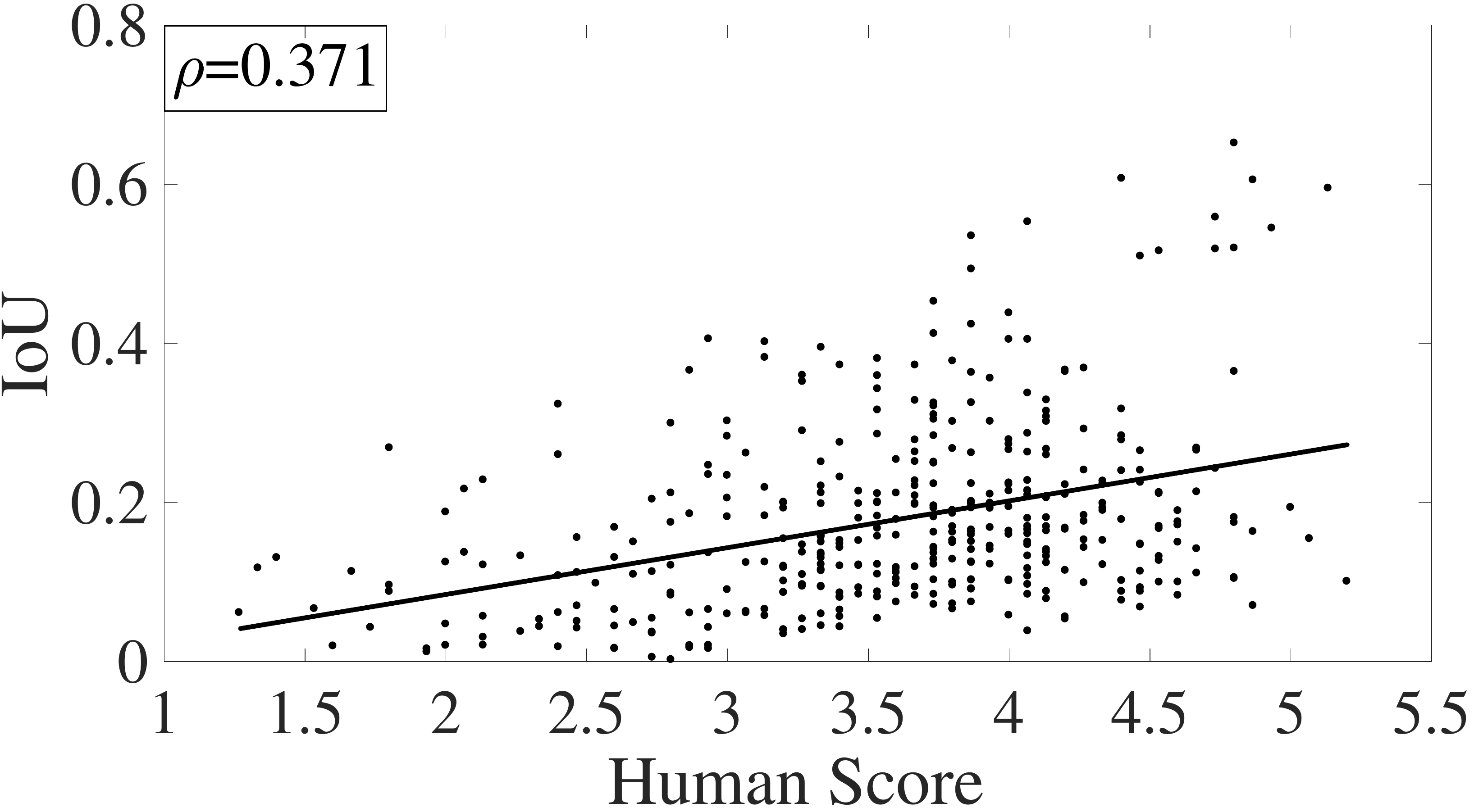}
\end{minipage}
\begin{minipage}[c]{0.33\linewidth}
    \centering
    \includegraphics[width=0.99\linewidth]{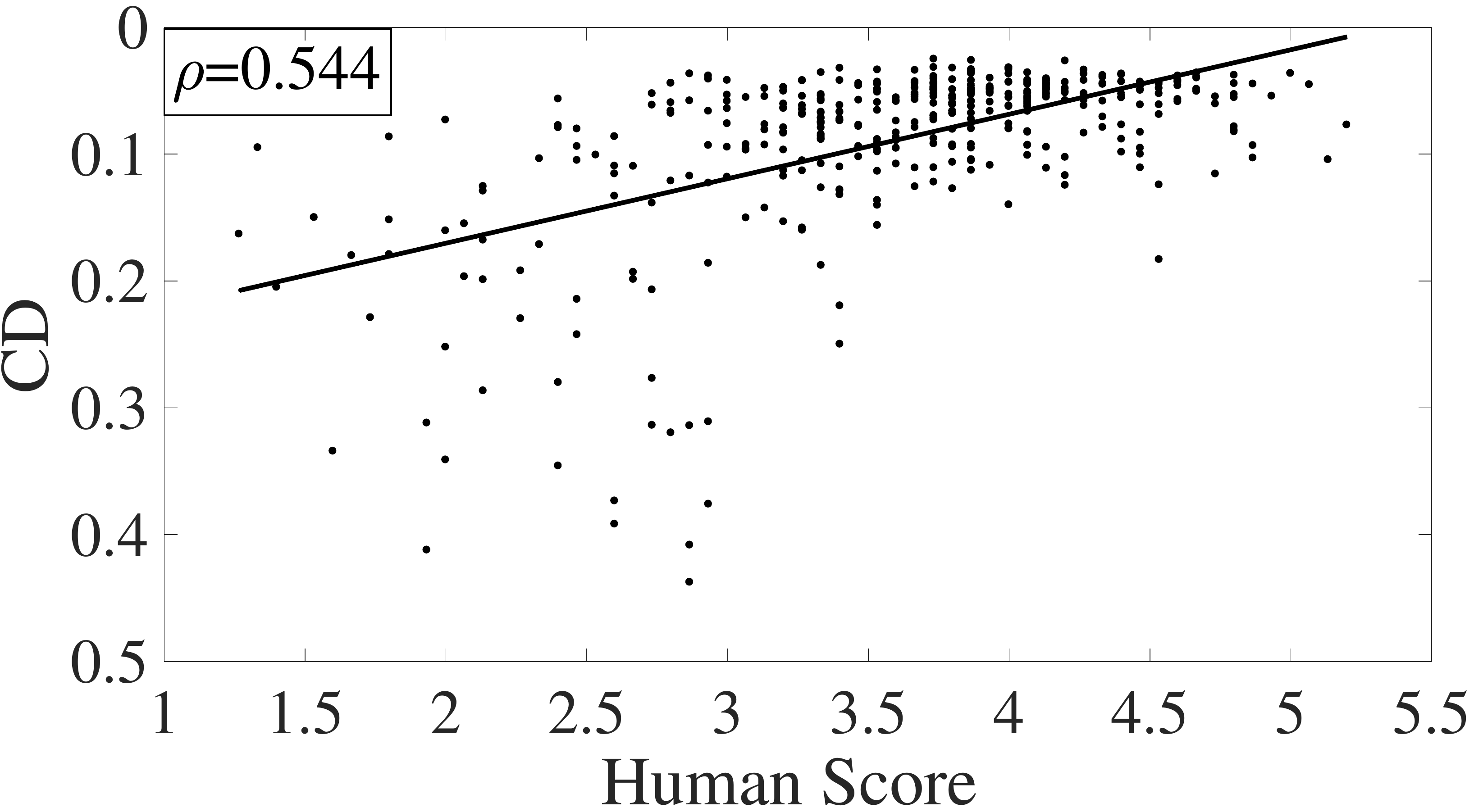}
\end{minipage}
\begin{minipage}[c]{0.33\linewidth}
    \centering
    \includegraphics[width=0.99\linewidth]{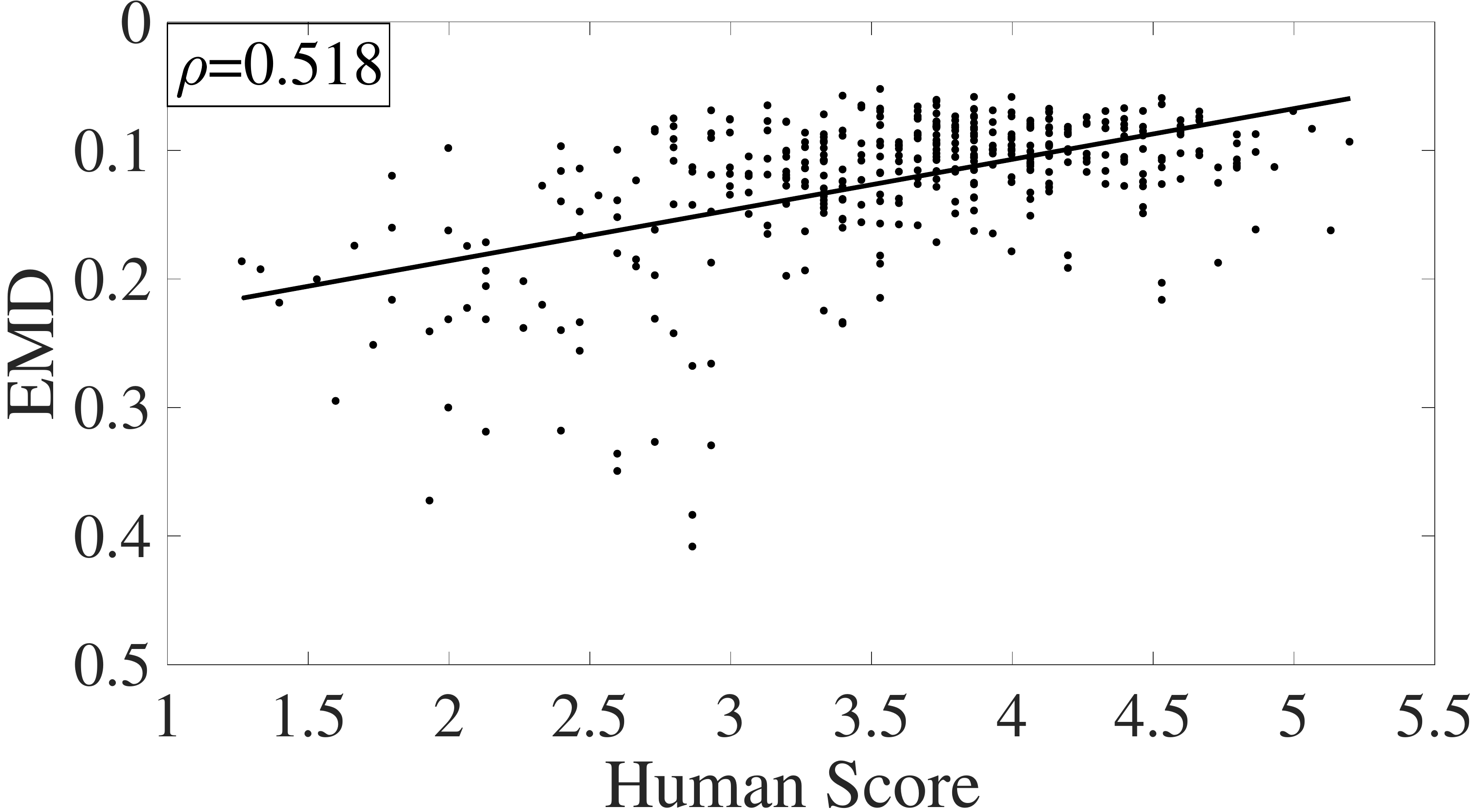}
\end{minipage}
\vspace{-5pt}
\caption{Scatter plots between humans' ratings of reconstructed shapes and their IoU, CD, and EMD. The three metrics have a Pearson's coefficient of 0.371, 0.544, and 0.518, respectively.} 
\label{fig:metric_scatter}
\vspace{-10pt}
\end{figure*}
\begin{figure}[t]
 	\centering
    \begin{tabular}{lcccc}
    \toprule
     & \multicolumn{2}{c}{Chairs} & \multicolumn{2}{c}{Sofas} \\
    \cmidrule(lr){2-3}\cmidrule(lr){4-5}
     & IoU & Match? & IoU & Match? \\
    \midrule
     PASCAL 3D+ \cite{Xiang2014PASCAL:} & 0.514 & 0.00 & 0.813 & 0.00 \\
     ObjectNet3D \cite{Xiang2016Objectnet3d:} & 0.570 & 0.16 & 0.773 & 0.08 \\
     IKEA \cite{Lim2013Parsing} & 0.748 & \textbf{1.00} & 0.918 & \textbf{1.00} \\
     \data (ours) & \textbf{0.835} & \textbf{1.00} & \textbf{0.926} & \textbf{1.00} \\
    \bottomrule
    \end{tabular}
    \captionof{table}{We compute the Intersection over Union (IoU) between manually annotated 2D masks and the 2D projections of 3D shapes. We also ask humans to judge whether the object in the images matches the provided shape.} 
    \label{tbl:analy_iou}
    \vspace{-15pt}
\end{figure}

\section{Metrics}
\label{sec:metric}

Designing a good evaluation metric is important to encourage researchers to design algorithms that reconstruct high-quality 3D geometry, rather than low-quality 3D reconstruction that overfits to a certain metric.

Many 3D reconstruction papers use Intersection over Union (IoU) to evaluate the similarity between ground truth and reconstructed 3D voxels, which may significantly deviate from human perception. In contrast, metrics like shortest distance and geodesic distance are more commonly used than IoU for matching meshes in graphics~\cite{kreavoy2007model,jain2006robust}. Here, we conduct behavioral studies to calibrate IoU, Chamfer distance (CD)~\cite{barrow1977parametric}, and Earth Mover's distance (EMD)~\cite{rubner2000earth} on how well they reflect human perception.

\subsection{Definitions}

The definition of IoU is straightforward. For Chamfer distance (CD) and Earth Mover's distance (EMD), we first convert voxels to point clouds, and then compute  CD and EMD between pairs of point clouds. 

\myparagraph{Voxels to a point cloud.} We first extract the isosurface of each predicted voxel using the Lewiner marching cubes~\cite{Lewiner03} algorithm. In practice, we use 0.1 as a universal surface value for extraction. We then uniformly sample points on the surface meshes and create the densely sampled point clouds. Finally, we randomly sample 1,024 points from each point cloud and normalize them into a unit cube for distance calculation. 

\myparagraph{Chamfer distance (CD).} The Chamfer distance (CD) between $S_1, S_2 \subseteq \mathbb{R}^3$ is defined as
\begin{equation}
\small
    \text{CD}(S_1, S_2) = \frac{1}{|S_1|} \sum_{x \in S_1} \min_{y \in S_2} \norm{x - y}_2 + \frac{1}{|S_2|}\sum_{y \in S_2} \min_{x \in S_1} \norm{x - y}_2.
\end{equation}
For each point in each cloud, CD finds the nearest point in the other point set, and sums the distances up. CD has been used in recent shape retrieval challenges~\cite{yi2017large}. 

\myparagraph{Earth Mover's distance (EMD).}  We follow the definition of EMD in Fan~\etal~\cite{fan2017point}. The Earth Mover's distance (EMD) between $S_1, S_2 \subseteq \mathbb{R}^3$ (of equal size, \ie, $|S_1| = |S_2|$) is
\begin{equation}
    \text{EMD}(S_1, S_2) = \frac{1}{|S_1|} \min_{\phi: S_1 \to S_2} \sum_{x \in S_1} ||x - \phi(x)||_2,
\end{equation}
where $\phi: S_1 \to S_2$ is a bijection. We divide EMD by the size of the point cloud for normalization. In practice, calculating the exact EMD value is computationally expensive; we instead use a $(1 + \epsilon)$ approximation algorithm~\cite{bertsekas1985distributed}. 

\subsection{Experiments}

We then conduct two user studies to compare these metrics and benchmark how they capture human perception.

\myparagraph{Which one looks better?}
We run three shape reconstructions algorithms (3D-R2N2~\cite{Choy20163d}, DRC~\cite{Tulsiani2017Multi}, and 3D-VAE-GAN~\cite{Wu2016Learning}) on 200 randomly selected images of chairs.  We then, for each image and every pair of its three constructions, ask three AMT workers to choose the one that looks closer to the object in the image. We also compute how each pair of objects rank in each metric. Finally, we calculate the Spearman's rank correlation coefficients between different metrics (\ie, IoU, EMD, CD, and human perception). \tbl{tbl:metric_corr} suggests that EMD and CD correlate better with human ratings.

\begin{figure}[t]
\centering
    \begin{tabular}{lcccc}
    \toprule
     & IoU & EMD & CD & Human \\
    \midrule
    IoU & 1 & 0.55 & 0.60 & 0.32 \\
    EMD & 0.55 & 1 & 0.78 & 0.43 \\
    CD & 0.60 & 0.78 & 1 & 0.49 \\
    Human & 0.32 & 0.43 & 0.49 & 1 \\
    \bottomrule
    \end{tabular}
    \vspace{-5pt}
    \captionof{table}{Spearman's rank correlation coefficients between different metrics. IoU, EMD, and CD have a correlation coefficient of 0.32, 0.43, and 0.49 with human judgments, respectively.}
    \vspace{-15pt}
    \label{tbl:metric_corr}
\end{figure}

\begin{figure*}[t]
\centering
\includegraphics[width=\linewidth]{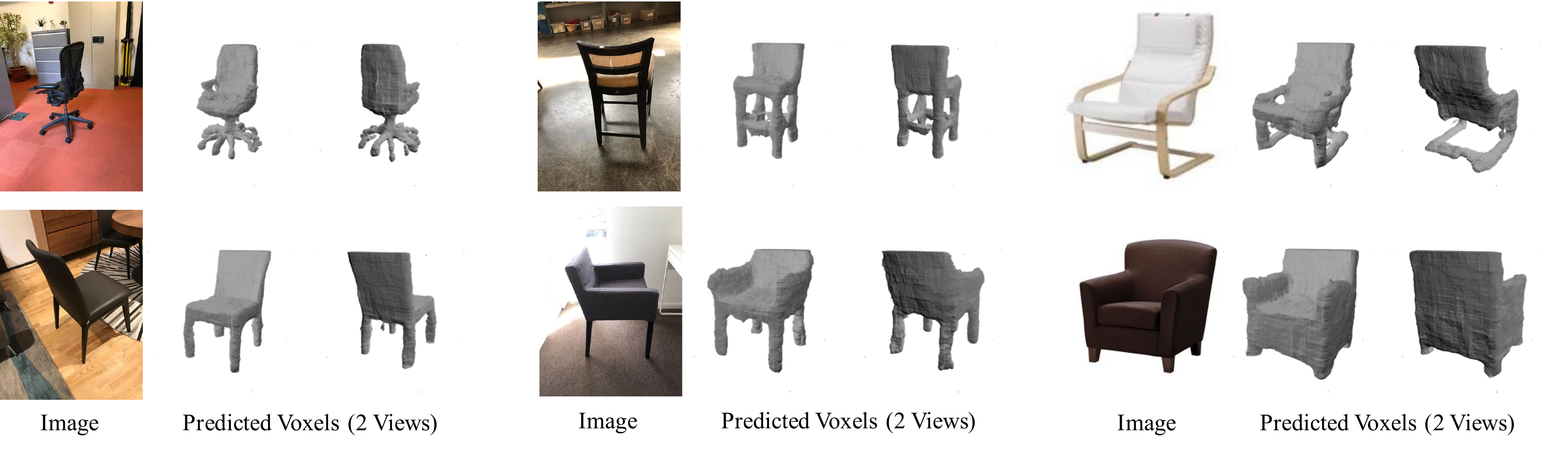}
\vspace{-20pt}
\caption{Results on 3D reconstructions of chairs. We show two views of the predicted voxels for each example.}
\vspace{-15pt}
\label{fig:eval_recon}
\end{figure*}

\myparagraph{How good is it?}

We randomly select 400 images, and show each of them to 15 AMT workers, together with the voxel prediction by DRC~\cite{Tulsiani2017Multi} and the ground truth shape. We then ask them to rate the reconstruction, on a scale of 1 to 7, based on how similar it is to the ground truth. The scatter plot in \fig{fig:metric_scatter} suggests that CD and EMD have higher Pearson's coefficients with human responses.

\begin{table}[t]
    \centering
 	\small
    \begin{tabular}{lccc}
    \toprule
     & IoU & EMD & CD \\
    \midrule
    3D-R2N2~\cite{Choy20163d} & 0.136 & 0.211 & 0.239 \\
    PSGN~\cite{fan2017point} & N/A & 0.216 & 0.200 \\
    3D-VAE-GAN~\cite{Wu2016Learning} & 0.171 & 0.176 & 0.182 \\
    DRC~\cite{Tulsiani2017Multi} & 0.265 & 0.144 & 0.160 \\
    MarrNet*~\cite{marrnet} & 0.231 & 0.136 & 0.144 \\
    AtlasNet~\cite{groueix2018} & N/A & 0.128 & 0.125 \\
    Ours (w/o Pose) & 0.267 & 0.124 & 0.124 \\
    Ours (w/ Pose) & \textbf{0.282} & \textbf{0.118} & \textbf{0.119} \\
    \bottomrule
    \end{tabular}
    \vspace{-5pt}
    \caption{Results on 3D shape reconstruction. Our model gets the highest IoU, EMD, and CD. We also compare our full model with a variant that does not have the view estimator. Results show that multi-task learning helps boost its performance. As MarrNet and PSGN predict viewer-centered shapes, while the other methods are object-centered, we rotate their reconstructions into the canonical view using ground truth pose annotations before evaluation.} 
    \vspace{-15pt}
    \label{tbl:eval_recon}
\end{table}

\section{Approach}
\label{sec:approach}

\data serves as a benchmark for shape modeling tasks including reconstruction, retrieval, and pose estimation. Here, we design a new model that simultaneously performs shape reconstruction and pose estimation, and evaluate it on \data. 

Our model is an extension of MarrNet~\cite{marrnet}, both of which use 2.5D sketches (the object's depth, surface normals, and silhouette) as an intermediate representation. It contains four modules: (1) a 2.5D sketch estimator that predicts the depth, surface normals, and silhouette of the object; (2) a 2.5D sketch encoder that encodes the 2.5D sketches into a low-dimensional latent vector; (3) a 3D shape decoder and (4) a view estimator that decodes a latent vector into a 3D shape and camera parameters, respectively. Different from MarrNet~\cite{marrnet}, our model has an additional branch for pose estimation. We briefly describe them below, and please refer to the supplementary material for more details.

\begin{figure*}[t]
\centering
\includegraphics[width=\linewidth]{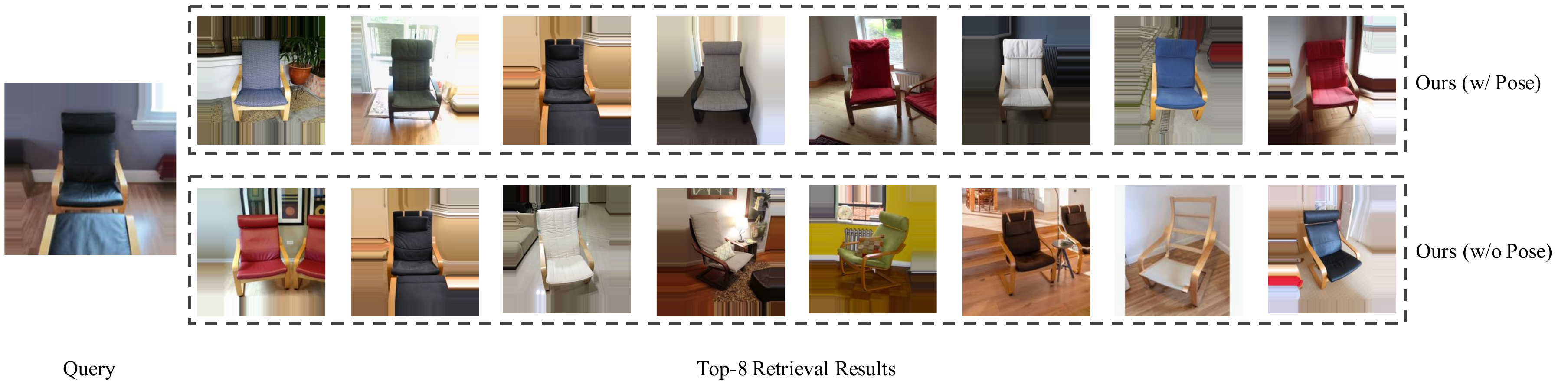}
\vspace{-20pt}
\caption{Results on shape retrieval. We show the top-8 retrieval results from our proposed method (with and without pose estimation). The variant with pose estimation tends to retrieve images of shapes in a similar pose.}
\vspace{-10pt}
\label{fig:eval_retri}
\end{figure*}
\begin{figure*}[t]
\centering
\includegraphics[width=\linewidth]{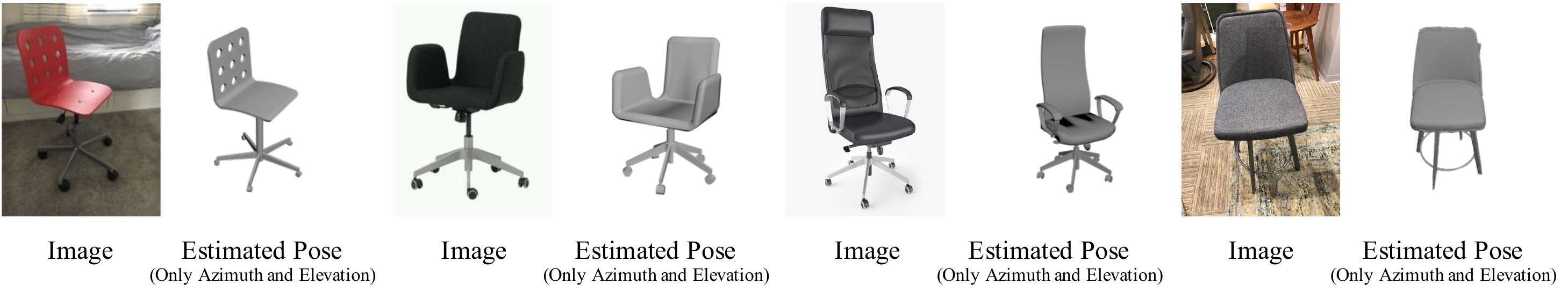}
\vspace{-20pt}
\caption{Results on pose estimation. Our method predicts azimuth and elevation accurately.}
\vspace{-15pt}
\label{fig:eval_pose}
\end{figure*}

\myparagraph{2.5D sketch estimator.}
The first module takes an RGB image as input and predicts the object's 2.5D sketches (its depth, surface normals, and silhouette). We use an encoder-decoder network. The encoder is based on a ResNet-18~\cite{He2015Deep} and turns a 256$\times$256 image into 384 feature maps of size 16$\times$16; the decoder has three branches for depth, surface normals, and silhouette, respectively, each consisting of four sets of 5$\times$5 transposed convolutional, batch normalization and ReLU layers, followed by one 5$\times$5 convolutional layer. All output sketches are of size 256$\times$256.

\myparagraph{2.5D sketch encoder.}
We use a modified ResNet-18~\cite{He2015Deep} that takes a four-channel image (three for surface normals and one for depth). Each channel is masked by the predicted silhouette. A final linear layer outputs a 200-D latent vector.

\myparagraph{3D shape decoder.}
Our 3D shape decoder has five sets of 4$\times$4$\times$4 transposed convolutional, batch-norm, and ReLU layers, followed by a 4$\times$4$\times$4 transposed convolutional layer. It outputs a voxelized shape of size 128$\times$128$\times$128 in the object's canonical view. 

\myparagraph{View estimator.}
The view estimator contains three sets of linear, batch normalization, and ReLU layers, followed by two parallel linear and softmax layers that predict the shape's azimuth and elevation, respectively. Here, we treat pose estimation as a classification problem, where the 360-degree azimuth angle is divided into 24 bins and the 180-degree elevation angle is divided into 12 bins.

\myparagraph{Training paradigm.}

For training, we use Mitsuba~\cite{Mitsuba} to render each chair in ShapeNet~\cite{Chang2015Shapenet:} from 20 random views using three types of backgrounds: 1/3 on a white background, 1/3 on high-dynamic-range backgrounds with illumination channels, and 1/3 on backgrounds randomly sampled from the SUN database~\cite{Xiao2010Sun}. We augment our training data by random color and light jittering. 

We first train the 2.5D sketch estimator. We then train the 2.5D sketch encoder and the 3D shape decoder (and the view estimator if we're predicting the pose) jointly. We finally concatenate them for prediction.

\section{Experiments}
\label{sec:exp}

We now evaluate our model and state-of-the-art algorithms on single-image 3D shape reconstruction, retrieval, and pose estimation, all using \data. For all experiments, we use the 2,894 untruncated and unoccluded chair images. 

\myparagraph{3D shape reconstruction.}

We compare our model, with and without the pose estimation branch, with the state-of-the-art systems, including 3D-VAE-GAN~\cite{Wu2016Learning}, 3D-R2N2~\cite{Choy20163d}, DRC~\cite{Tulsiani2017Multi}, and MarrNet~\cite{marrnet}. We use pre-trained models offered by the authors and we crop the input images as required by each algorithm. The results are shown in \tbl{tbl:eval_recon} and \fig{fig:eval_recon}. Our model outperforms the state-of-the-arts in all metrics. Our full model gets better results compared with the variant without the view estimator, suggesting multi-task learning helps to boost its performance. Also note the discrepancy among metrics: MarrNet has a lower IoU than DRC, but according to EMD and CD, it performs better.

\begin{table}[t]
 	\centering
 	\setlength{\tabcolsep}{2pt}
    \begin{tabular}{lcccccc}
    \toprule
     & R@1 & R@2 & R@4 & R@8 & R@16 & R@32\\
    \midrule
    3D-VAE-GAN \cite{Wu2016Learning} & 0.02 & 0.03 & 0.07 & 0.12 & 0.21 & 0.34 \\
    MarrNet \cite{marrnet} & 0.42 & 0.51 & 0.57 & 0.64 & 0.71 & 0.78 \\
    Ours (w/ Pose) & 0.42 & 0.48 & 0.55 & 0.63 & 0.70 & 0.76 \\
    Ours (w/o Pose) & \textbf{0.53} & \textbf{0.62} & \textbf{0.71} & \textbf{0.78} & \textbf{0.85} & \textbf{0.90} \\
    \bottomrule
    \end{tabular}
    \vspace{-5pt}
    \caption{Results on image-based shape retrieval, where R@K stands for Recall@K. Our model (without the pose estimation module) achieves the highest numbers. Our model (with the pose estimation module) does not perform as well, because it sometimes retrieves images of objects with the same pose, but not exactly the same shape.} 
    \label{tbl:metric_retr}
    \vspace{-15pt}
\end{table}

\myparagraph{Image-based, fine-grained shape retrieval.}

For shape retrieval, we compare our model with 3D-VAE-GAN~\cite{Wu2016Learning} and MarrNet~\cite{marrnet}. We use the latent vector from each algorithm as its embedding of the input image, and use L2 distance for image retrieval. For each test image, we retrieve its K nearest neighbors from the test set, and use Recall@K~\cite{Jegou2011Product} to compute how many retrieved images are actually depicting the same shape. Here we do not consider images whose shape is not captured by any other images in the test set. The results are shown in \tbl{tbl:metric_retr} and \fig{fig:eval_retri}. Our model (without the pose estimation module) achieves the highest numbers; our model (with the pose estimation module) does not perform as well, because it sometimes retrieves images of objects with the same pose, but not exactly the same shape. 

\begin{table}[t]
 	\centering
    \setlength{\tabcolsep}{3pt}
    \begin{tabular}{lccccccc}
    \toprule
    & \multicolumn{4}{c}{Azimuth} & \multicolumn{3}{c}{Elevation} \\
    \cmidrule(lr){2-5}\cmidrule(lr){6-8}
    \# of views & 4 & 8 & 12 & 24 & 4 & 6 & 12 \\
    \midrule
    Render for CNN & 0.71 & 0.63 & 0.56 & 0.40 & 0.57 & 0.56 & 0.37 \\
    Ours & \textbf{0.76} & \textbf{0.73} & \textbf{0.61} & \textbf{0.49} & \textbf{0.87} & \textbf{0.70} & \textbf{0.61} \\
    \bottomrule
    \end{tabular}
    \vspace{-5pt}
    \caption{Results on 3D pose estimation. Our model outperforms Render for CNN~\cite{Su2015Render} in both azimuth and elevation.}
    \vspace{-15pt}
    \label{tbl:eval_pose}
\end{table}

\myparagraph{3D pose estimation.}

We compare our method with Render for CNN~\cite{Su2015Render}. We calculate the classification accuracy for both azimuth and elevation, where the azimuth is divided into 24 bins and the elevation into 12 bins. \tbl{tbl:eval_pose} suggests that our model outperforms Render for CNN in pose estimation. Qualitative results are included in \fig{fig:eval_pose}.

\section{Conclusion}

We have presented \data, a large-scale dataset of well-aligned 2D images and 3D shapes. We have also explored how three commonly used metrics correspond to human perception through two behavioral studies and proposed a new model that simultaneously performs shape reconstruction and pose estimation. Experiments showed that our model achieved state-of-the-art performance on 3D reconstruction, shape retrieval, and pose estimation. We hope our paper will inspire future research in single-image 3D shape modeling.
 \myparagraph{Acknowledgements.} This work is supported by NSF \#1212849 and \#1447476, ONR MURI N00014-16-1-2007, the Center for Brain, Minds and Machines (NSF STC award CCF-1231216), the Toyota Research Institute, and Shell Research. J. Wu is supported by a Facebook fellowship. 

{\small
\bibliographystyle{ieee}
\bibliography{3ddata}
}

\begin{figure*}[t]
\centering
\includegraphics[width=\linewidth]{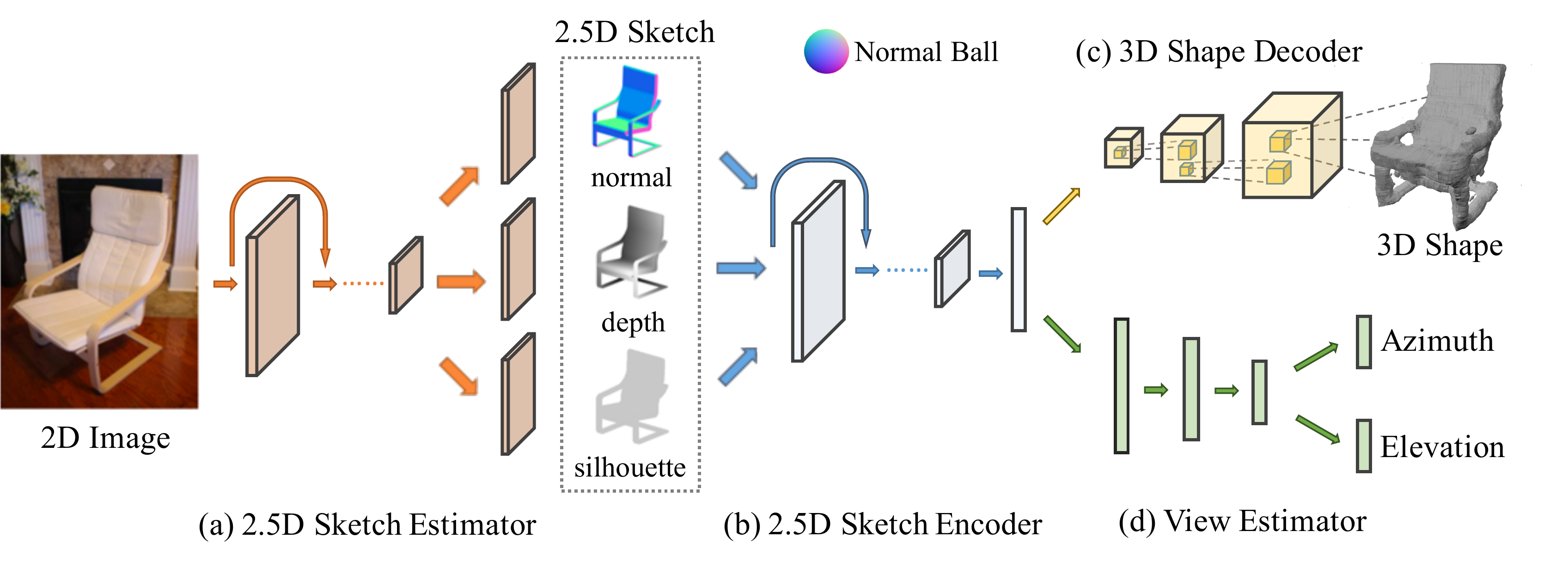}
\vspace{-20pt}
\caption{Our model has four major components: (a) a 2.5D sketch estimator, (b) a 2.5D sketch encoder, (c) a 3D shape decoder, and (d) a view estimator. Our model first predicts 2.5D sketches from an RGB image. It then encodes the 2.5D sketches into a latent vector. Finally, a 3D shape is decoded from the 3D shape decoder, and azimuth and elevation are estimated by the view estimator.}
\label{fig:method_pip}
\vspace{-5pt}
\end{figure*}
\begin{table*}[t]
 	\centering
    \begin{tabular}{cccc}
    \toprule
    \multicolumn{3}{c}{Type} & Configurations\\
    \midrule
    \multicolumn{3}{c}{ResNet-18~\cite{He2015Deep}} & without the last two layers (avg pool and fc) \\
    \multicolumn{3}{c}{deconv} & \#maps: 512 to 384, kernel: 5$\times$5, stride: 2, padding: 2 \\
    \multicolumn{3}{c}{batchnorm} & - \\
    \multicolumn{3}{c}{relu} & - \\
    deconv & deconv & deconv & \#maps: 384 to 384, kernel: 5$\times$5, stride: 2, padding: 2 \\
    batchnorm & batchnorm & batchnorm & - \\
    relu & relu & relu & - \\
    deconv & deconv & deconv & \#maps: 384 to 384, kernel: 5$\times$5, stride: 2, padding: 2 \\
    batchnorm & batchnorm & batchnorm & - \\
    relu & relu & relu & - \\
    deconv & deconv & deconv & \#maps: 384 to 192, kernel: 5$\times$5, stride: 2, padding: 2 \\
    batchnorm & batchnorm & batchnorm & - \\
    relu & relu & relu & - \\
    deconv & deconv & deconv & \#maps: 192 to 96, kernel: 5$\times$5, stride: 2, padding: 2 \\
    batchnorm & batchnorm & batchnorm & - \\
    relu & relu & relu & - \\
    conv & conv & conv & \#maps: 96 to 3 (for normal) / to 1 (for others), kernel: 5$\times$5, stride: 1, padding: 2 \\
    \bottomrule
    \end{tabular}
    \caption{The architecture of our 2.5D sketch estimator}
    \label{tbl:net_12}
\end{table*}

\newpage
\renewcommand\thesection{\Alph{section}}
\setcounter{section}{0} 

\section{Network Parameters}
\label{sec:net_para}

As mentioned in Section 6 in the main text, we proposed a new model that simultaneously performs 3D shape reconstruction and camera view estimation. Here we provide more details about the network structure.

As shown in \fig{fig:method_pip}, our model consists of four components: (1) a 2.5D sketch estimator, which estimates 2.5D sketches from an RGB image, (2) a 2.5D sketch encoder, which encodes 2.5D sketches into a 200-dimensional latent vector, (3) a 3D shape decoder, which decodes a latent vector into voxels and (4) a view estimator, which estimates the camera view from a latent vector.

\myparagraph{2.5D sketch estimator. }
\tbl{tbl:net_12} shows the network configuration summary of the 2.5D sketch estimator. We use an encoder-decoder network. The first four rows in \tbl{tbl:net_12} shows the encoder's structure and the other rows describe the decoder. The encoder takes in an input RGB image of size 256$\times$256 and encodes it into 384 16$\times$16 feature maps. The decoder takes in 384 16$\times$16 feature maps and decodes them into the object's surface normals, depth, and silhouette of size 256$\times$256.

For the encoder, we use a truncated ResNet-18~\cite{He2015Deep} with last two layers (average pooling and fully connected) removed. The truncated ResNet-18 is followed by a transposed convolutional layer, a batch normalization layer, and a ReLU layer. For the decoder, we use four sets of 5$\times$5 transposed convolutional layers, batch normalization layers and ReLU layers, followed by one 5$\times$5 convolutional layer. We do not share weights of layers between three sketches (\ie, surface normal, depth, silhouette).

\myparagraph{2.5D sketch encoder. }
The 2.5D sketch encoder is modified from a ResNet-18. It takes in a four-channel image with size 256$\times$256 obtained by stacking the three-channel surface normal image and single-channel depth image, both of which are masked by the silhouette. It then encodes them into a 200-dimensional latent vector.

For the first layer of ResNet-18, we change the number of input channels from 3 to 4. We also change the average pooling layer into an adaptive average pooling layer. For the last fully connected layer, we change the output dimensional to 200.

\myparagraph{3D shape decoder. }
\tbl{tbl:net_3} shows the network architecture of the 3D shape decoder. It takes in a 200-dimensional latent vector and decodes it into a voxel grid of size 128$\times$128$\times$128. We use five sets of 4$\times$4$\times$4 3D transposed convolutional layers, 3D batch normalization layers and ReLU layers, followed by one 4$\times$4$\times$4 transposed convolutional layer.

\myparagraph{View estimator. }
\tbl{tbl:net_4} shows the network configuration summary of the view estimator. We use three sets of fully connected, batch normalization, and ReLU layers, followed by two parallel fully connected and softmax layers that predict azimuth and elevation, respectively.

	\begin{table}[t]
         	\centering
            \begin{tabular}{cc}
            \toprule
            Type & Configurations\\
            \midrule
            deconv3d & \#maps:200 to 512, k:4$\times$4$\times$4, s:1, p:0 \\
            batchnorm3d & - \\
            relu & - \\
            deconv3d & \#maps:512 to 256, k:4$\times$4$\times$4, s:2, p:1 \\
            batchnorm3d & - \\
            relu & - \\
            deconv3d & \#maps:256 to 128, k:4$\times$4$\times$4, s:2, p:1 \\
            batchnorm3d & - \\
            relu & - \\
            deconv3d & \#maps:128 to 64, k:4$\times$4$\times$4, s:2, p:1 \\
            batchnorm3d & - \\
            relu & - \\
            deconv3d & \#maps:64 to 32, k:4$\times$4$\times$4, s:2, p:1 \\
            batchnorm3d & - \\
            relu & - \\
            deconv3d & \#maps:32 to 1, k:4$\times$4$\times$4, s:2, p:1 \\
            \bottomrule
            \end{tabular}
            \caption{The architecture of our 3D shape decoder. k, s, p stand for kernel size, stride and padding size respectively.}
            \label{tbl:net_3}
        \end{table}
    	\begin{table}[t]
         	\centering
         	\setlength{\tabcolsep}{1pt}
            \begin{tabular}{ccc}
            \toprule
            \multicolumn{2}{c}{Type} & Configurations\\
            \midrule
            \multicolumn{2}{c}{fc} & 200 to 800 \\
            \multicolumn{2}{c}{batchnorm1d} & - \\
            \multicolumn{2}{c}{relu} & - \\
            \multicolumn{2}{c}{fc} & 800 to 400 \\
            \multicolumn{2}{c}{batchnorm1d} & - \\
            \multicolumn{2}{c}{relu} & - \\
            \multicolumn{2}{c}{fc} & 400 to 200 \\
            \multicolumn{2}{c}{batchnorm1d} & - \\
            \multicolumn{2}{c}{relu} & - \\
            fc & fc & 200 to 24 (for azimuth) / to 12 (for elevation) \\
            softmax & softmax & - \\
            \bottomrule
            \end{tabular}
            \caption{The architecture of our view estimator}
            \label{tbl:net_4}
        \end{table}
\begin{figure*}[t!]
\centering
\includegraphics[width=\linewidth]{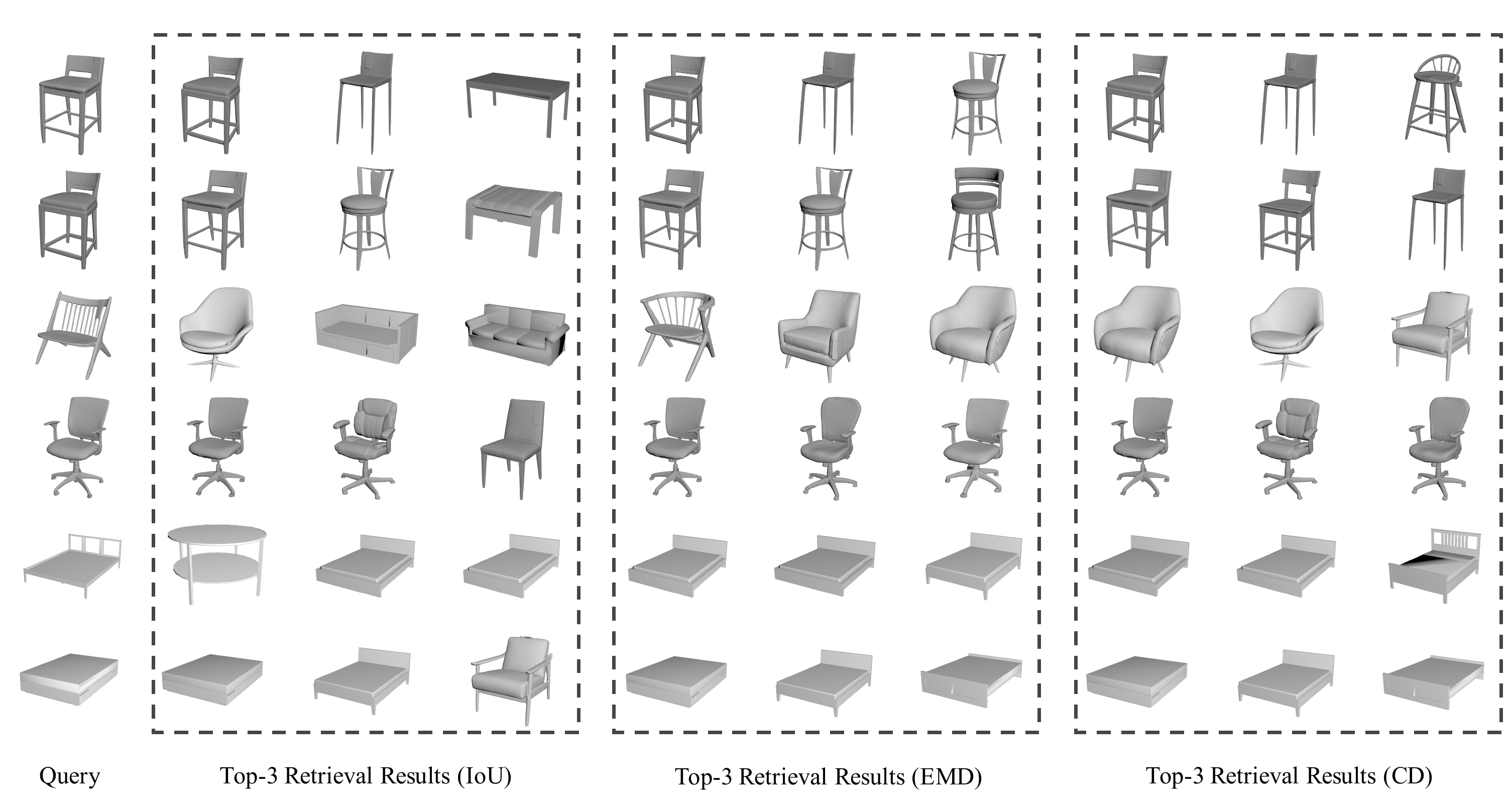}
\caption{Three nearest neighbors retrieved from \data using different metrics. EMD and CD work slightly better than IoU.}
\label{fig:metric}
\end{figure*}

\section{Training Paradigms}
\label{sec:train_para}

As mentioned in Section 7 in the main text, we train our proposed method and test it on three different tasks. Here we provide more details about training.

We first train the 2.5D sketch estimator. We then train the 2.5D sketch encoder and the 3D shape decoder (and the view estimator if we're predicting the pose) jointly.

\myparagraph{2.5D sketch estimation. }
The loss function is defined as the sum of mean squared error between predicted sketches and ground truth sketches (with size average). Specifically,
\begin{align}
    \text{loss}_1 & = \text{MSE}(\text{depth}_\text{pred}, \text{depth}_\text{gt}) + \text{MSE}(\text{normal}_\text{pred}, \text{normal}_\text{gt}) \nonumber \\ & + \text{MSE}(\text{silhouette}_\text{pred}, \text{silhouette}_\text{gt}),
\end{align}
where MSE is mean square error with size average, pred stands for prediction, and gt stands for ground truth.

The batch size is 4. We use Adam~\cite{Kingma2015Adam:} as the optimizer and set the learning rate to $2\times10^{-4}$. The model is trained for 270 epochs, each with 6,000 batches. We choose to use the one with the minimum validation loss.

\myparagraph{Shape and view estimation. }
The loss function is defined as the weighted sum of the 3D reconstruction loss and the pose estimation loss. The loss function for 3D reconstruction is 
\begin{equation}
    \text{loss}_\text{recon} = \text{BCE}_{L}(\text{voxel}_\text{pred}, \text{voxel}_\text{gt}),
\end{equation}
where $\text{BCE}_{L}$ is the binary cross-entropy between the target and the output logits (no sigmoid applied) with size average, pred stands for prediction, and gt stands for ground truth.
The loss function for pose estimation is 
\begin{align}
    \text{loss}_\text{pose} & = \text{BCE}(\text{azimuth}_\text{pred}, \text{azimuth}_\text{gt}) \nonumber \\
    & + \text{BCE}(\text{elevation}_\text{pred}, \text{elevation}_\text{gt}),
\end{align}
where $\text{BCE}$ is the binary cross-entropy between the target and the output with size average, pred stands for prediction, and gt stands for ground truth. Note that we have already applied softmax to azimuth and elevation predictions in our model. The global loss function is thus
\begin{equation}
    \text{loss}_2 = \text{loss}_\text{recon} + \alpha \cdot \text{loss}_\text{pose}
\end{equation}

We set $\alpha$ to 0.6. The batch size is 4. We use stochastic gradient descent with a momentum of 0.9 as the optimizer and set the learning rate to 0.1. The model is trained for 300 epochs, each with 6,000 batches. We choose to use the one with the minimum validation loss.
\section{Evaluation Metrics}
\label{sec:eval}

Here, we explain in detail our evaluation protocol for single-image 3D shape reconstruction. As different voxelization methods may result in objects of different scales in the voxel grid, for a fair comparison, we preprocess all voxels and point clouds before calculating IoU, CD and EMD. 

For IoU, we first find the bounding box of the object with a threshold of 0.1, pad the bounding box into a cube, and then use trilinear interpolation to resample to the desired resolution ($\text{32}^\text{3}$). Some algorithms reconstruct shapes at a resolution of $\text{128}^\text{3}$. In this case, we first, apply a 4$\times$ max pooling before trilinear interpolation; without the max pooling, the sampling grid can be too sparse and some thin structure can be left out. After the resampling of both the output voxel and the ground truth voxel, we search for an optimal threshold that maximizes the average IoU score over all objects, from 0.01 to 0.50 with a step size of 0.01. 

For CD and EMD, we first sample a point cloud from the voxelized reconstructions. For each shape, we compute its isosurface with a threshold of 0.1, and then sample 1,024 points from the surface. All point clouds are then translated and scaled such that the bounding box of the point cloud is centered at the origin with its longest side being 1. We then compute CD and EMD for each pair of point clouds. 

\section{Nearest Neighbors of 3D Shapes}
\label{sec:metric}

In Section 5 in the main text, we have compared three different metrics from two different perspectives. We here compare them in another way: for a 3D shape, we retrieve three nearest neighbors from \data according to IoU, EMD and CD, respectively. Results are shown in \fig{fig:metric}. EMD and CD perform slightly better than IoU.

\section{Sample Data Points in \data}
\label{sec:data}

\begin{figure*}[p]
\centering
\vspace{-12pt}
\includegraphics[width=\linewidth]{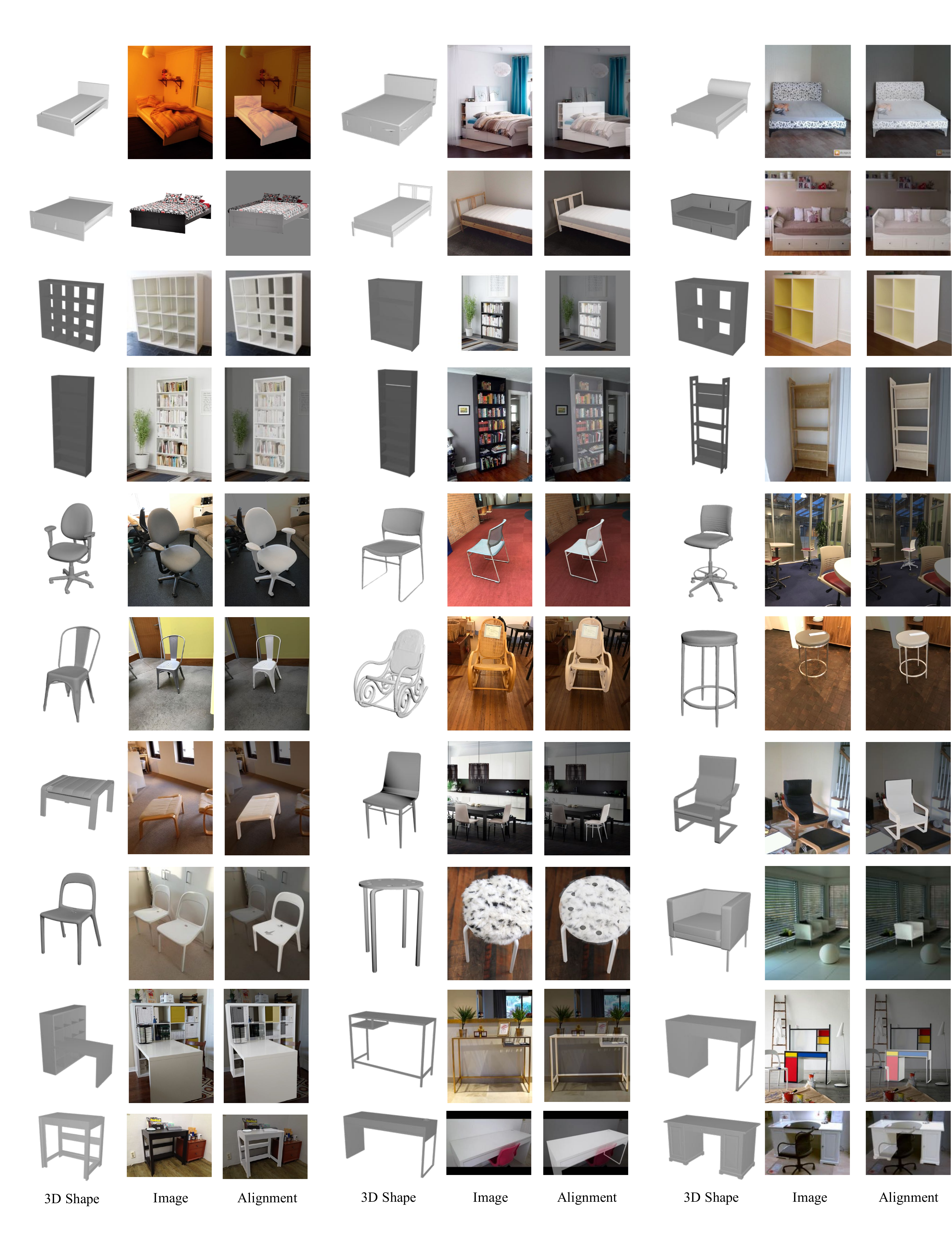}
\caption{Sample images and corresponding shapes in \data. From left to right: 3D shapes, 2D images, 2D-3D alignment. The \nth{1} and \nth{2} rows are beds, the \nth{3} and \nth{4} rows are book selves, the \nth{5} and \nth{6} rows are scanned chairs, the \nth{7} and \nth{8} rows are chairs whose 3D shapes come from IKEA~\cite{Lim2013Parsing}, and the \nth{9} and \nth{10} rows are desks.}
\label{fig:data_sample_1}
\end{figure*}

\begin{figure*}[p]
\centering
\vspace{-7pt}
\includegraphics[width=\linewidth]{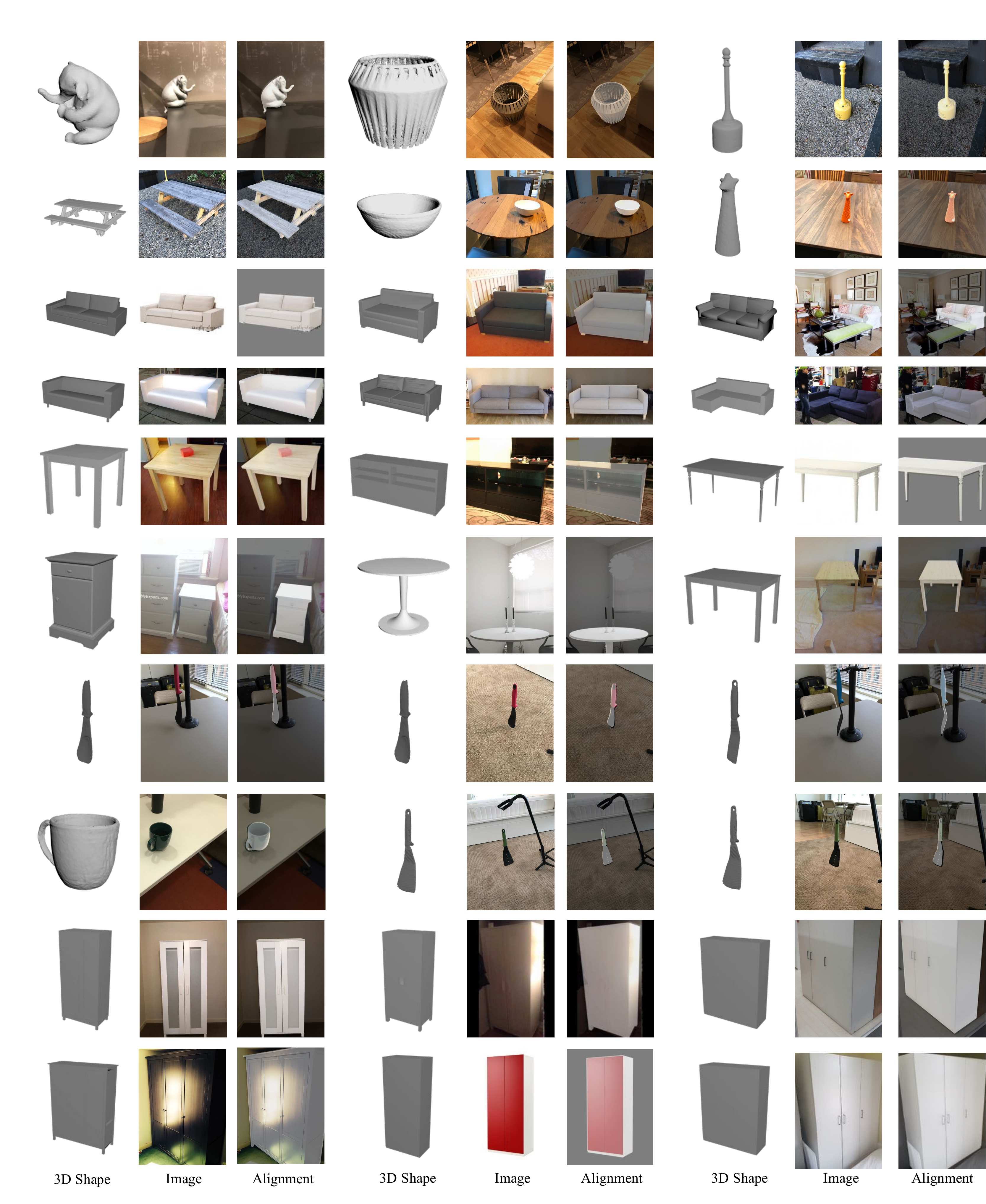}
\caption{Sample images and corresponding shapes in \data. From left to right: 3D shapes, 2D image, 2D-3D alignment. The \nth{1} and \nth{2} rows are miscellaneous objects, the \nth{3} and \nth{4} rows are sofas, the \nth{5} and \nth{6} rows are tables, the \nth{7} and \nth{8} rows are tools, and the \nth{9} and \nth{10} rows are wardrobes.}
\label{fig:data_sample_2}
\end{figure*}

\begin{figure*}[t]
\centering
\includegraphics[width=\linewidth]{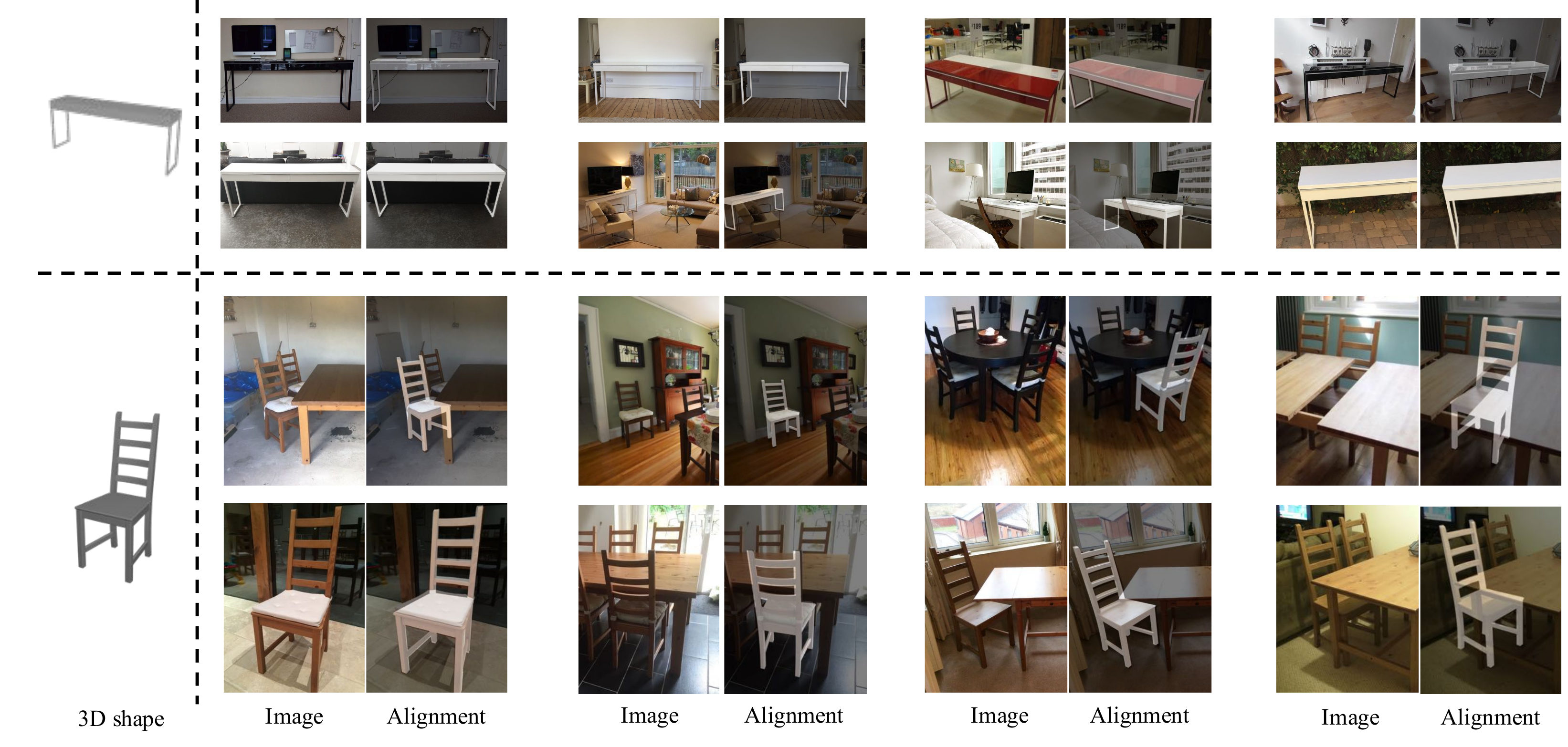}
\caption{Sample images and corresponding shapes in \data. The two 3D shapes are each associated with a diverse set of 2D images.}
\label{fig:same_shape}
\end{figure*}

We supply more sample data points in \figs{fig:data_sample_1}, \ref{fig:data_sample_2}, and \ref{fig:same_shape}. \figs{fig:data_sample_1} and \ref{fig:data_sample_2} show the diversity of 3D shapes and the quality of 2D-3D alignment in \data. \fig{fig:same_shape} shows that each shape in \data is matched with a rich set of 2D images.

\end{document}